\documentclass[10pt,twocolumn,letterpaper]{article}

\usepackage{iccv}
\usepackage{times}
\usepackage{epsfig}
\usepackage{graphicx}
\usepackage{amsmath}
\usepackage{amssymb}
\usepackage{subfigure}
\usepackage{booktabs} 
\usepackage{color}

\usepackage[breaklinks=true,bookmarks=false]{hyperref}

\iccvfinalcopy 

\def\iccvPaperID{****} 

\begin{document}

\title{ARGAN: Attentive Recurrent Generative Adversarial Network for Shadow Detection and Removal}

\author{Bin Ding\textsuperscript{1}, 
Chengjiang Long\textsuperscript{2}\thanks{This work was co-supervised by Chengjiang Long and Chunxia Xiao.}, 
Ling Zhang\textsuperscript{3,1}, 
Chunxia Xiao\textsuperscript{1} \\
\textsuperscript{1}{School of Computer Science, Wuhan University, Wuhan, Hubei, China} \\
\textsuperscript{2}{Kitware Inc., Clifton Park, NY, USA} \\
\textsuperscript{3}{Wuhan University of Science and Technology, Wuhan, Hubei, China} \\
{\tt\small dingbin@whu.edu.cn, chengjiang.long@kitware.com, czhling@wust.edu.cn, cxxiao@whu.edu.cn}
}

\maketitle

\begin{abstract}
  In this paper we propose an attentive recurrent generative adversarial network (ARGAN) to detect and remove shadows in an image. 
The generator consists of multiple progressive steps. At each step a shadow attention detector is firstly exploited to generate an attention map which specifies shadow regions in the input image.Given the attention map, {{a negative residual by a shadow remover encoder}} will recover a shadow-lighter or even a shadow-free image. A discriminator is designed to classify whether the output image in the last progressive step is real or fake. Moreover, ARGAN is suitable to be trained with a semi-supervised strategy to make full use of sufficient unsupervised data. The experiments on four public datasets have demonstrated that our ARGAN is robust to detect both simple and complex shadows and to produce more realistic shadow removal results. It outperforms the state-of-the-art methods, especially in detail of recovering shadow areas.
\end{abstract}

\section{Introduction}

Shadow exists in most images and is formulated in interaction {{among}} the illumination, object materials and scene geometry. Obviously, the detected shadows can provide important clues for various applications of visual scene understanding, such as scene geometry depiction~\cite{Okabe2009Attached}, camera location~\cite{junejo2008estimating}, object relighting~\cite{Karsch2014Automatic}, and scene illumination inference~\cite{Han2016StackGAN}. Meanwhile, the shadow removal is able to boost the performance of some computer vision and computer graphics tasks, such as object detection and tracking~\cite{Miki2000Moving, long2014accurate, luo2019end}, object recognition~\cite{Cucchiara2002Improving, hua2013collaborative, long2015multi, Long_2017_CVPR, hua2018collaborative}, intrinsic image decomposition~\cite{Li2018Learning}. Therefore, it is desirable to develop an effective method of shadow detection and removal.

 \begin{figure}[t]
  \centering
  {\includegraphics[width=0.97\linewidth]{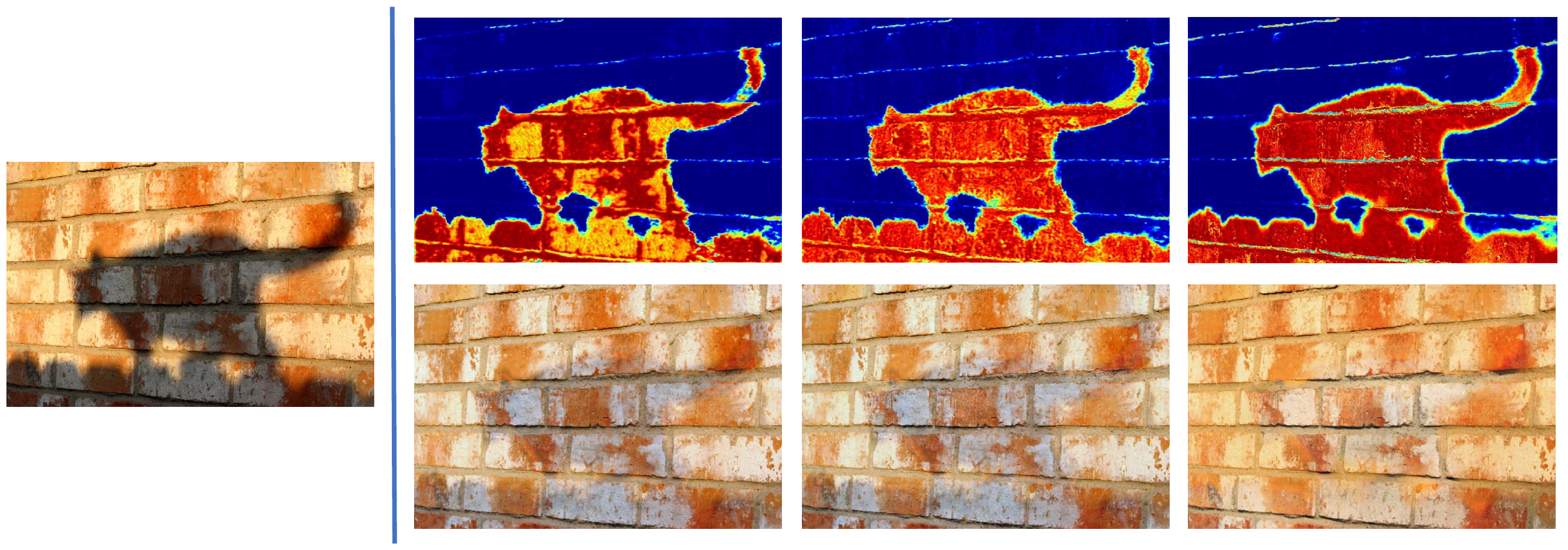}}
  \caption{Given an input image (left) with shadow, our goal is to generate a more accurate attention map gradually indicating the detected shadow region  (right-top) and recover a more realistic shadow-removal image (right-bottom) gradually with multiple progressive steps.}
\label{fig0}
\end{figure}


Previous works for shadow detection and removal can be mainly divided into two categories. One category is traditional methods~\cite{guo2011single, zhang2015shadow} based on some prior knowledge, such as consistent illumination in shadow regions. Its shortcoming is that the priors might dissatisfy some shadow images under a complex environment so that the performance of shadow removal result cannot be guaranteed. The other is deep learning methods~\cite{qu2017deshadownet, hu2018direction,wang2018stacked,le2018a,zhu2018bidirectional} whose effectiveness rely heavily on supervised data to learn a robust model. Particularly, however, when the training data is insufficient, such deep learning methods often appear color distortion or other problems in the shadow removal results.

\begin{figure*}[t]
  \centering
  {\includegraphics[width=0.97\linewidth]{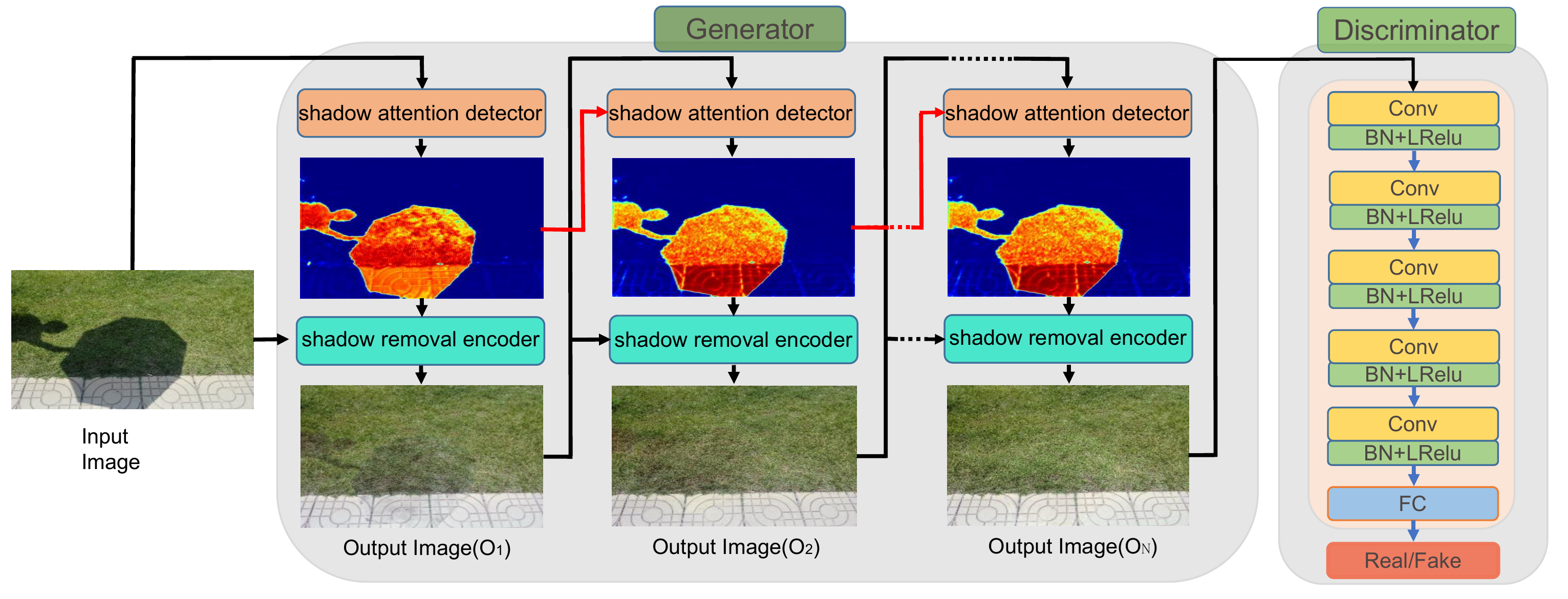}}
  \caption{The framework overview of the proposed ARGAN consists of two components, {\em i.e.}, a generator and a discriminator. The generator consists of $N$ shadow attention detectors and $N$ shadow image encoders. Each shadow attention detector is designed to generate the shadow attention map, and each shadow removal encoder is to produce shadow-lighter or even shadow-free image. The discriminator is formed by five convolutional layers and  a fully connected layer to classify the output shadow-free image as real or fake.}
\label{fig1}
\end{figure*}

In this paper, we propose a novel attentive recurrent generative adversarial network (ARGAN) for shadow detection and removal. As illustrated in Figure~\ref{fig1}, the generator involves multiple progressive step for shadow detection and removal in a coarse-to-fine fashion, and the discriminator is designed to classify whether the generated shadow-removal image at the last step from the generator is real or fake. At each progressive step in the generator, a shadow attention detector is used to generate an attention map. Then a shadow removal encoder is designed to combine the previous shadow-removal image and the current detected shadow attention map to obtain a negative residual~\cite{fu2017removing} for recovering the shadow-lighter or even shadow-free images.

The intuition behind multiple progressive steps in the generator is that it is much easier to detect and remove shadow gradually with a step-by-step approach so that we are able to handle shadows especially with complicated scenes. The detected shadow region and recovered shadow-lighter image from the previous step are the input of the present step. Therefore {\color{red}{,}} we are able to explore a recurrent unit such as Long and Short Term Memory (LSTM) \cite{hochreiter1997long} to reserve the valuable and detailed information to make sure that the detected shadow regions are more and more accurate, and that the shadow-removal images are more and more realistic, as illustrated in Figure~\ref{fig0}. 


We shall emphasize that we adopt the adversarial training process~\cite{goodfellow2014generative} between the generator and the discriminator to generate a shadow-removal image. With the number of epoches increases, both models improve their functionalities so that it becomes harder and harder to distinguish a generated shadow-removal image from a real shadow-free image. Therefore, after a certain large number of training epochs, we can utilize the learned parameters in the generator to generate a shadow attention map and a shadow-removal image at each progressive step. The output from the last step is our final result.

Moreover, we apply a semi-supervised learning strategy~\cite{simo2018mastering} to make full use of sufficient unsupervised shadow images available online by modifying the original adversarial loss to cover both labeled data and unlabeled data. {{We}} can first use the generator and generate a shadow-removal image for any input images with shadow, and then just use the discriminator to discriminate whether the generated image is real or not. This treatment can improve the generalization ability and robustness of our ARGAN.

Several aspects distinguish our work from the previous shadow detection and removal methods~\cite{guo2011single, zhang2015shadow, qu2017deshadownet, hu2018direction,wang2018stacked,le2018a,zhu2018bidirectional}. First of all, our proposed ARGAN adopts adversarial training process to optimize each shadow attention detector and each shadow removal encoder in the generator. Secondly, the generator involves multiple progressive steps for shadow detection and removal in a coarse-to-fine fashion so that it can handle shadows with complicated environment. Thirdly, a semi-supervised strategy by incorporating sufficient unsupervised shadow images available online is able to increase the robustness of our network. We evaluate our proposed ARGAN on the four public datasets and compare with the state-of-the-art methods on the performance of both shadow detection and shadow removal. The results clearly demonstrate the efficacy of our proposed model.



\section{Related Work}
The related work can be divided into four categories: {\em shadow detection methods}, {\em shadow removal methods}, {\em Generative Adversarial Network} and {\em attention mechanism}.

\textbf{Shadow detection methods} involve traditional methods, using user interactions~\cite{Gryka2015Learning, zhang2015shadow, Eli2011Shadow} and hand-crafted features~\cite{Lalonde2010Detecting, guo2011single, vicente2018leave}, and recent deep learning methods~\cite{khan2016automatic, Vicente2016Large, Nguyen2017Shadow, Hu2017Direction, le2018a, zhu2018bidirectional} for automatic shadow detection. To specify, Khan \etal~\cite{khan2016automatic} detected the shadow by combining the boundary and region ConvNets in the CRF model. Vicente \etal~\cite{Vicente2016Large} proposed a semantic-aware patch level CNN architecture for shadow detection. Nguyen \etal~\cite{Nguyen2017Shadow} detected shadow using conditional generative adversarial networks. Hu \etal~\cite{Hu2017Direction} detected shadow by analyzing image context in a direction-aware manner. However, these methods only work well on image with simple shadow. They cannot detect accurate shadow with complex scenes.

\textbf{Shadow removal methods} based on gradient domain manipulation \cite{Mohan2007Editing, Feng2008Texture},  illumination~\cite{zhang2015shadow, xiao2013fast, khan2016automatic, shor2008shadow}, color transferring~\cite{reinhard2001color}, accurate shadow matte~\cite{chuang2003shadow, Gryka2015Learning, wu2007natural} and depth information~\cite{xiao2014shadow} have been exposed for a long time. Recently three deep learning methods have been proposed for shadow removal. One is Qu \etal's multi-context embedding network~\cite{qu2017deshadownet} integrating high-level semantic context for shadow removal. {{One}} is Hu \etal's ~\cite{hu2018direction} using direction-aware spatial context features for shadow detection and removal. {{Another}} is Wang~\etal's GAN-based method~\cite{wang2018stacked} which jointly learns shadow detection and shadow removal. Different from~\cite{wang2018stacked}, our proposed ARGAN involves multiple progressive steps with attentive recurrent units in the generator to achieve better performance on shadow removal.

\textbf{Generative Adversarial Network} (GAN)~\cite{goodfellow2014generative} and  its variants~\cite{Han2016StackGAN} have been proposed to deal with various image-to-image translation problems, such as image super-resolution \cite{ledig2017photo}, image inpainting~\cite{pathak2016context}, style transfer~\cite{li2016precomputed} and domain adaptation/transfer \cite{liu2017unsupervised, isola2017image, tran2019disentangling}, raindrop removal~\cite{qian2018attentive}, shadow detection and removal \cite{wang2018stacked}. 
Unlike~\cite{qian2018attentive} which only progressively updates the attention map with the same input image for a one-step removal, our proposed ARGAN progressively detects shadow and removals shadow step by step in a coarse-to-fine fashion.

\textbf{Attention mechanism}~\cite{bahdanau2014neural} is designed to encode sequence data based on the assigned importance score of each element, which has attained significant improvement in various tasks in natural language processing~\cite{kiela2018dynamic, tao2019radical}, speech recognition~\cite{chorowski2015attention}, computer vision~\cite{zhang2018progressive}, image captioning~\cite{xu2015show, lu2017knowing, gao2019hierarchical} and visual question answering (VQA)~\cite{lu2016hierarchical, anderson2018bottom}. 
Different from~\cite{zhang2018progressive} which uses the progressive and recurrent idea to integrate multiple contextual information of multi-level features, our ARGAN progressively and recurrently updates the shadow attention maps and the shadow removal images in the generator in a coarse-to-fine fashion so that it can handle shadows with complicated environment.


\section{Approach}

As illustrated in Figure~\ref{fig1}, we present a attentive recurrent generative adversarial network (ARGAN) to explore the mapping relationship from shadow images to the corresponding shadow-free images. Like all the generative adversarial networks, our ARGAN contains two components, {\em i.e.}, a generator to produce a shadow-free image as real as possible, and a discriminator to classify whether the generated shadow-free image is indeed a real image or not. 

At the generative stage, given an input shadow image $I$, we iteratively update the detected shadow region indicated with an attention map $A_i$ by shadow attention detector $G_{det}^i$ and output a shadow-lighter or even shadow-free image $O_i$ by shadow removal encoder $G_{rem}^i$ at $i$-th step by the following equations:
\begin{equation}
\label{eqn:01a}
A_i=\left\{\begin{matrix}
G_{det}^i(I)\quad\quad\quad\quad& i=1\\
G_{det}^i(O_{i-1}, A_{i-1})& i > 1
\end{matrix}\right.
\end{equation}
\begin{equation}
\label{eqn:01b}
O_i=\left\{\begin{matrix}
G_{rem}^i(I, A_i)\quad\quad& i=1\\
G_{rem}^i(O_{i-1}, A_i)& i > 1
\end{matrix}\right.
\end{equation}

At the discriminative stage, we design a discriminator $D$ to encode the final output shadow-free image $O_N$ with semi-supervised strategy to handle the supervised data (the ground-truth shadow-free images $F$) and unsupervised data ($F$ is missing) under the adversarial framework.

In the following subsections, we are going to discuss the generative network, the discriminative network, and the loss functions, as well as the implementation details.

\subsection{Generative Network}
Our generative network is composed of $N$ progressive steps and each step has one shadow attention detector and one shadow removal encoder.

\textbf{Shadow attention detector.} We incorporate attention mechanism to selectively choose what our network wants to observe, locate shadows of the input image and make the attention of shadow removal encoder focusing on the detected shadow regions. As shown in Figure~\ref{fig1}, a recurrent unit through Long and Short Term Memory (LSTM) \cite{hochreiter1997long} can be formulated into our recurrent attention network, in which LSTM can make full use of the intermediate output of the previous step in the recurrent network and, as a prior, generate attention map which represents shadow region in the later steps.
\begin{figure}
  \centering
  \includegraphics[width=0.90\linewidth]{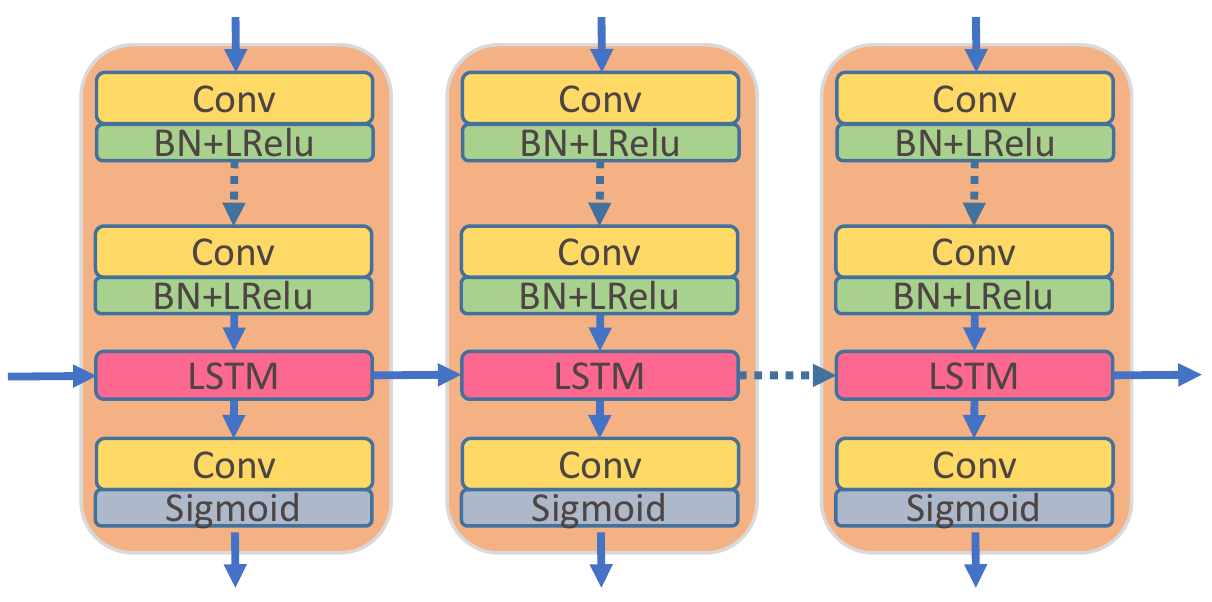}
  \caption{Each shadow attention detector consists of ten convolutional layers with 64 as output channel number, one LSTM layer and one convolutional layer. Note that every convolutional layer here is followed by a batch normalization and Leaky-ReLU active function. The stride is 1.}
\label{fig2a}
\end{figure}

As illustrated in Figure~\ref{fig2a}, the shadow attention detector in our recurrent network is designed as ten convolutional layers with batch normalization and Leaky ReLU activation function (Conv+BN+LRelu) 
to extract feature from the input image. A convolutional LSTM unit~\cite{xingjian2015convolutional} is to estimate the shadow regions by combining information at the previous step and passing the estimated result to the next step and a convolutional layer to generate a shadow attention map, which is the shadow matte in Equation~\ref{eqn:01a}.

\begin{figure}
  \centering
  \includegraphics[width=0.24\linewidth]{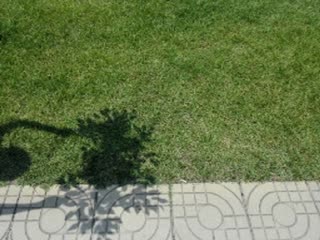}
  \includegraphics[width=0.24\linewidth]{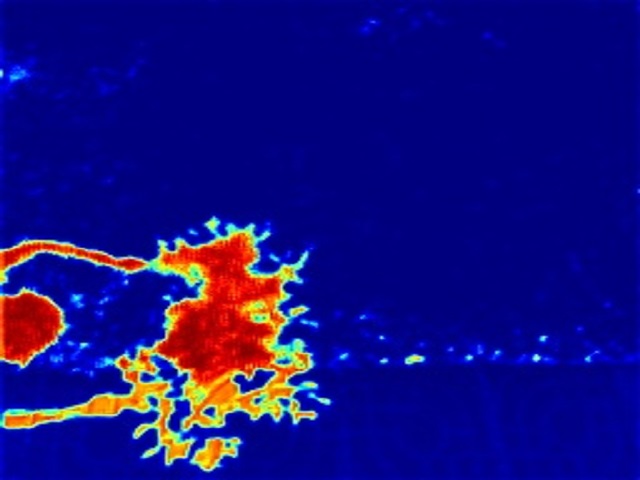}
  \includegraphics[width=0.24\linewidth]{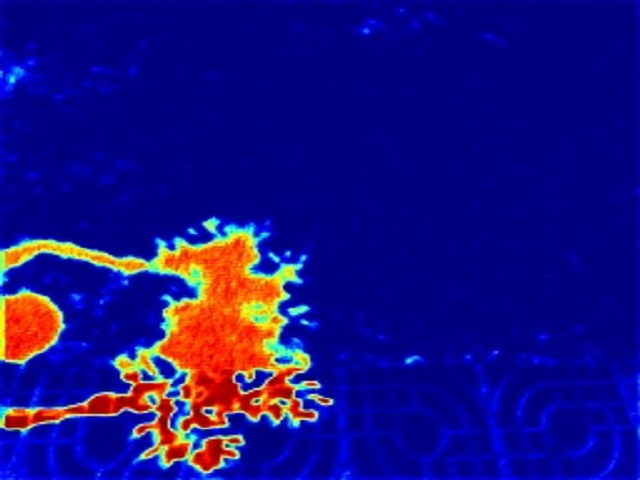}
  \includegraphics[width=0.24\linewidth]{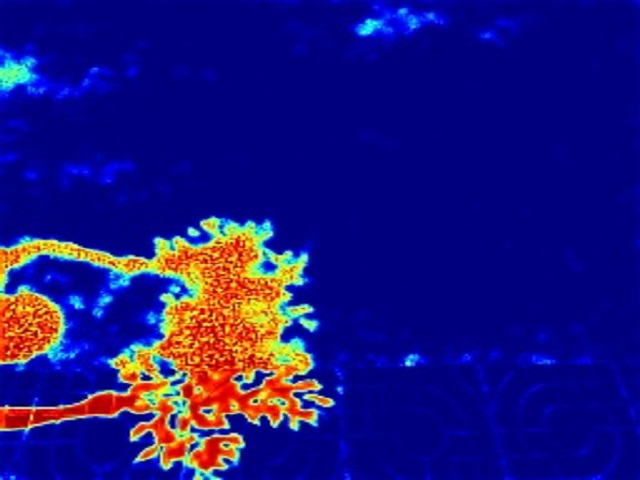}\\
  \caption{Attention maps generated by shadow attention detector at the first three progressive iterations. From left to right are input shadow image, region attention maps $A_1$, $A_2$ and $A_3$, respectively.}
\label{fig3a}
\end{figure}

Note that all the $N$ shadow attention detectors share the same architecture. Each output of attention map $A_i$ is a matrix. Each value in such a matrix is in the range from 0 to 1, rather than a binary mask. The larger the value, the more attention at this area. It indicates that the area of large value is more likely to be marked as shadow region. As shown in Figure~\ref{fig3a}, the red area with greater attention value close to 1 is more likely to be shadow region, whereas the blue area where the attention value is close to 0 is more likely to be considered as non-shadow region. In this way, our attention map can effectively distinguish soft shadow and hard shadow in the image by giving different attention values.


We observe that the attention module may not focus on all shadow regions of the image initially. The attention will constantly and gradually cover to the target region in the subsequent recurrent iterations. Therefore, it gradually shifts the attention to focus on all shadow regions. Figure~\ref{fig3a} shows one example of attention maps generated at the first three progressive steps. The detected shadow region marked in red are more and more accurate as the step number increases.

\textbf{Shadow removal encoder.} In order to remove shadow in an image, we follow the idea of negative residual~\cite{fu2017removing} and design the shadow removal encoder, which incorporates both the image with shadow, and the detected shadow regions attention to generate a negative residual for recovering a shadow-lighter or shadow-free image. 

\begin{figure}
  \centering
  \includegraphics[width=0.999\linewidth]{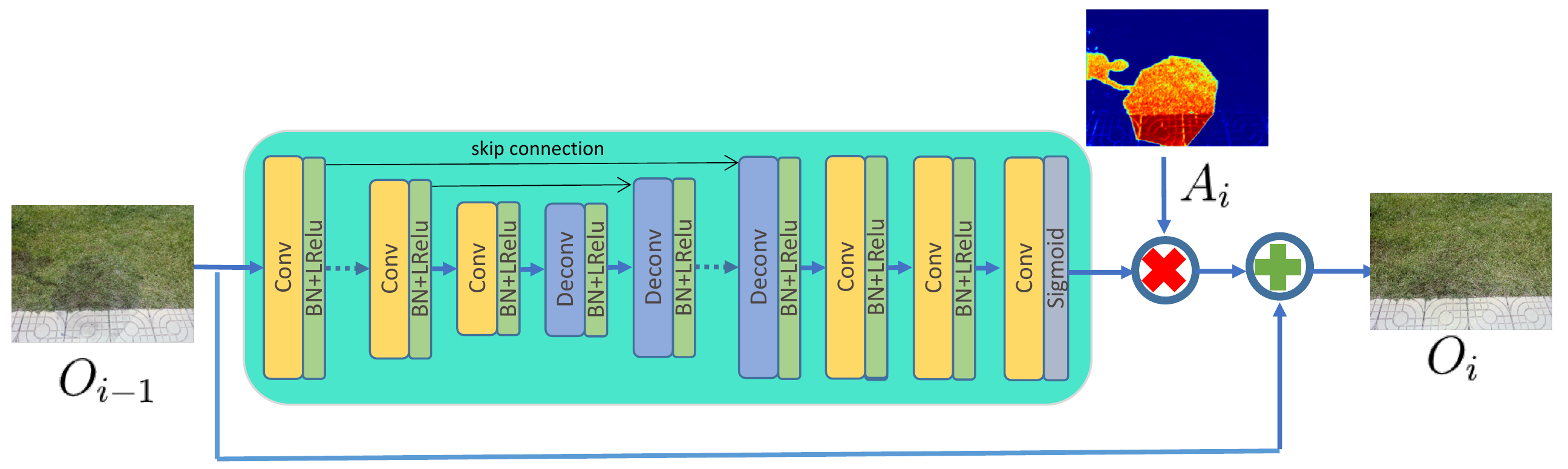}
  \caption{The aritecture of the shadow removal encoder. It consists of 8 Conv+BN+LRelu and 8 Deconv+BN+LRelu. The skip connections are linked between convolutional layers and deconvolutional layers. After that, 2 Conv+BN+LRelu, 1 Conv and 1 Sigmoid layer generates a corresponding map with 3 channels and the same size of the input $O_{i-1}$. A product operation is applied between the sigmoid output and the detected attention map $A_i$. Then a negative residual is obtained to recover a shadow-lighter or even shadow-free image $O_i$.  }
\label{fig2b}
\end{figure}

As illustrated in Figure~\ref{fig2b}, the encoder firstly uses eight Conv+BN+LRelu to extract feature from the image. Then it takes eight deconvolutional layers with batch normalization and Leaky ReLU activation function (Deconv+BN+LRelu) to generate image with feature data of a particular distribution. Skip connection \cite{he2016deep} is applied between convolutional layers and deconvolutional layers because it is able to not only increase the number of channels in the network, but also preserve the context information of front layer.

Following the last deconvolutional layer, 2 Conv+BN+LRelu are applied to extract the feature map and a convolutional layer with sigmoid activation to convert the feature map into a corresponding map with 3 channels and the same size of the input $O_{i-1}$. The parameters for all the convolutional layers and deconvolutional layers are summarized in Table~\ref{tab0}. Finally, we convert the sigmoid output to the residual information with the detected attention map $A_i$ by a product operation to get a negative residual for recovering a shadow-lighter or even a shadow-free image $O_i$ from the input image $O_{i-1}$.


\newcommand{\tabincell}[2]{\begin{tabular}{@{}#1@{}}#2\end{tabular}}
\begin{table}[]
\centering

\scriptsize
\begin{tabular}{|p{6.0mm}|p{6.0mm}|p{5.5mm}|p{5.0mm}|p{7.0mm}|p{5.5mm}|p{5.5mm}|p{5.0mm}|}
\hline
Layer & \tabincell{c}{Output}  & Kernel & Stride  & Layer & \tabincell{c}{Output} & Kernel & Stride \\ \hline\hline
Conv & 64  & $3\times3$ & 2  & Deconv & 512 & $4\times4$ & 2  \\ \hline
Conv & 128  & $3\times3$ & 2  & Deconv & 512 & $4\times4$ & 2  \\ \hline
Conv & 256  & $3\times3$ & 2  & Deconv & 512 & $4\times4$ & 2  \\ \hline
Conv & 512  & $3\times3$ & 2  & Deconv & 256 & $4\times4$ & 2  \\ \hline
Conv & 512  & $3\times3$ & 2  & Deconv & 128 & $4\times4$ & 2  \\ \hline
Conv & 512  & $3\times3$ & 2  & Deconv & 64 & $4\times4$ & 2  \\ \hline
Conv & 512  & $3\times3$ & 2  & Deconv & 3 & $4\times4$ & 2  \\ \hline
Conv & 512  & $3\times3$ & 2  & Conv & 3 & $3\times3$ & 1  \\ \hline
Deconv & 512  & $4\times4$ & 2  & Conv & 3 & $3\times3$ & 1  \\ \hline
\end{tabular}
\vspace{4pt}
\caption{The architecture for shadow removal encoder. Conv means the convolutional layer, Deconv means the deconvolutional layer. Output channels denotes for the amount of output channels in current layer. Kernel means the convolutional kernel size. Stride denotes the moving step size of convolutional kernel.} \label{tab0}
\end{table}

\begin{figure}
  \centering
  \includegraphics[width=0.24\linewidth]{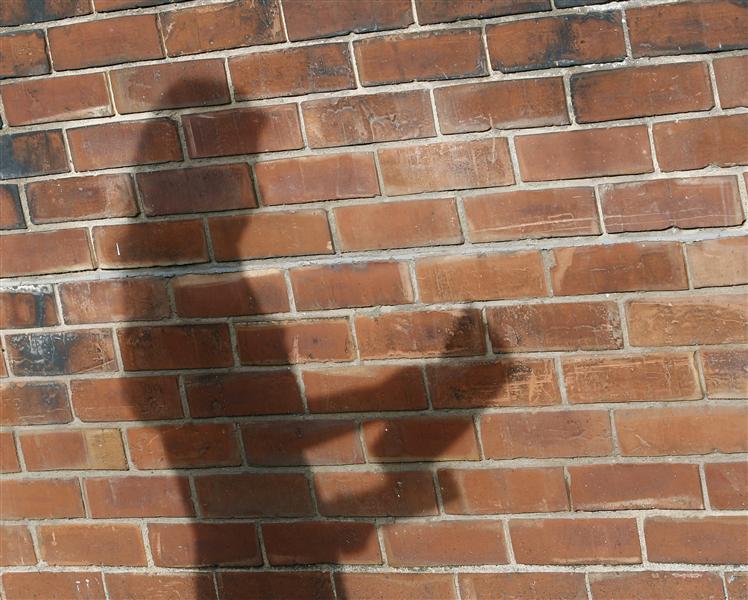}
  \includegraphics[width=0.24\linewidth]{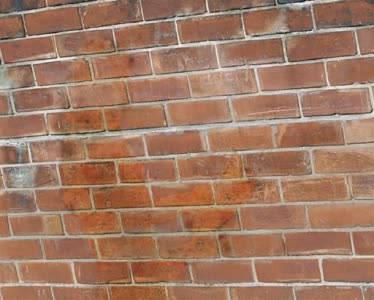}
  \includegraphics[width=0.24\linewidth]{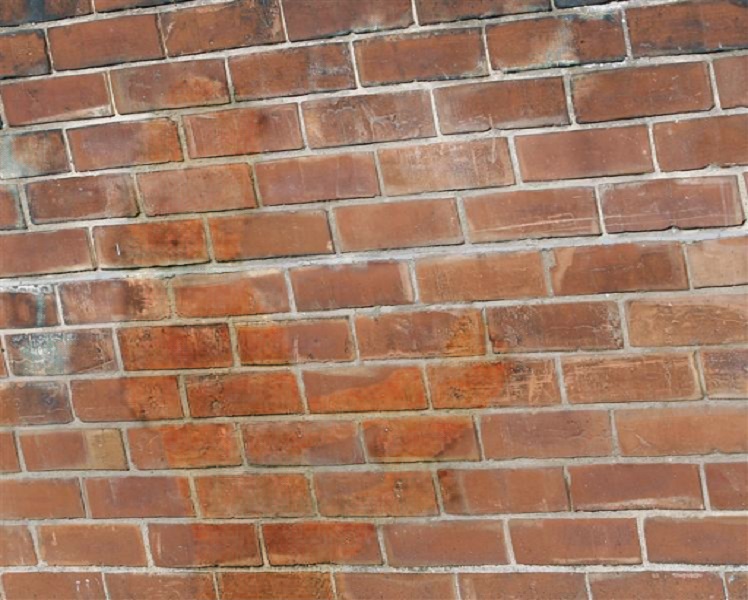}
  \includegraphics[width=0.24\linewidth]{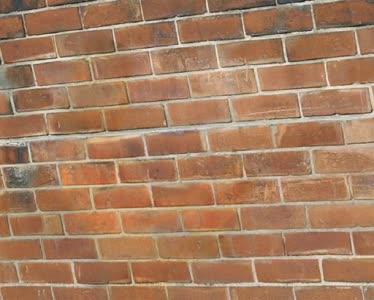}\\
  \caption{Output images generated by shadow removal encoder at the first three progressive iterations. From left to right are input shadow image, output images $O_1$, $O_2$ and $O_3$, respectively.}
\label{fig3b}
\end{figure}

As we can observe in Figure~\ref{fig3b}, as the progressive step number increases, the shadow in the output image becomes lighter and lighter, and the last output image $O_3$ is almost shadow-free.


\subsection{Discriminative Network}
Discriminator is designed as a binary classifier to predict whether the final output image $O_N$ from the generator is real or fake. It is worth mentioning that both generator and discriminator constantly improve their ability. Finally, they achieve a balanced state that the image produced by generator seems to be a real shadow-free image, which is consistent with our expectation that our generator model can produce a realistic shadow-free image so that the discriminator may consider it as a real shadow-free image.


To make it simple, we design our discriminator with five Conv+BN+LRelu and one fully connected layer, as shown in Figure~\ref{fig1}. The output channel numbers for all these six layer are 64, 128, 256, 512 and 1, respectively. Note that the kernel size for convolutional layers is $4\times4$, and the stride length is 2. The last fully connected layer outputs the actual probability value for the input image. The fake image and the real image are distinguished by calculating the cross entropy loss between them.

It worths paying attention to that some shadow scenes may be missed in the shadow datasets. To solve this problem, inspired by~\cite{simo2018mastering}, we apply semi-supervised learning strategy to our network. We use shadow image without ground-truth as unsupervised data in the training process. For each training process, we also input an unsupervised data to the generator and generate a shadow-free image. The discriminator discriminates whether the generated image is real or not. The semi-supervised strategy can improve the generalization ability of our network and make our generator models more robust.

In addition, to make our ARGAN more stable, we use the latest spectral normalization \cite{miyato2018spectral} method to stabilize the training process of discriminator network, because spectral normalization is a simple and effective standardized method for limiting the optimization process of the discriminator in GAN, and it can make the whole Generator model better.

\subsection{Loss functions}
The loss function we use to optimize our ARGAN comes from the shadow attention detector, shadow removal encoder, and discriminator.
The total loss $L_{total}$  can be formulated as:
\begin{equation}
\label{eqn:02}
L_{total}=L_{det}+L_{rem}+L_{adv},
\end{equation}
where the corresponding loss components are described as following.

\textbf{Shadow attention detector loss $L_{det}$} in each shadow attention detector is defined as the mean square error (MSE) between shadow matte $M$ (which is obtained by comparing the ground-truth shadow-free image and its corresponding shadow image) and the output attention map. In our generator, we apply the shadow attention detector for $N$ iterations, and the loss function is expressed as:
\begin{equation}
\label{eqn:02}
L_{det}=\sum_{i=1}^{N}\beta_i V_{\text{MSE}}(A_i,M),
\end{equation}
where $\beta_i$ is the weight of the MSE loss at $i$-th iteration, $\beta_i={0.7}^{N-i+1}$, and $V_{\text{MSE}}(A_i,M)$ is the mean square error between $M$ and $A_i$.

\textbf{Shadow removal encoder loss $L_{rem}$} contains accuracy loss and perceptual loss \cite{johnson2016perceptual}. We define the loss function as:
\begin{equation}
\label{eqn:03}
L_{rem}=\sum_{i=1}^{N}L_{mse}(O_i, F)+\sum_{i=1}^{N}L_{per}(O_i, F),
\end{equation}
where $O_i$ is the shadow-lighter or even shadow-free image generated by shadow removal encoder, $F$ is the corresponding ground-truth shadow-free image, $L_{mse}(O_i,F)$ is accuracy loss, and $L_{per}(O_i,F)$ is perceptual loss.

$L_{mse}(O_i,F)$ is used to measure the difference between the ground-truth image and shadow-free image generated by shadow removal encoder at the $i$-th iteration. The smaller the value of MSE, the more accurate the shadow removal encoder. The accuracy loss function is defined as:
\begin{equation}
\label{eqn:05}
L_{mse}(O_i,F)= \beta_i V_{\text{MSE}}(O_i, F),
\end{equation}

$L_{per}(O_i,F)$ is used to calculate global difference between the ground-truth image and the shadow removal result. We extract image features using the pre-trained VGG16 model \cite{simonyan2014very} on the ImageNet dataset. The loss function is defined as :
  \begin{equation}
\label{eqn:06}
L_{per}(O_i,F)= V_{MSE}(\text{VGG}(O_i),\text{VGG}(F)),
\end{equation}
where $\text{VGG}(O_i)$ and $\text{VGG}(F)$ are the feature of image $O_i$ and $F$ extracted from VGG16 model.

\textbf{Adversarial loss $L_{adv}$} with supervised learning
is expressed as:
\begin{equation}
\begin{split}
\label{eqn:04a}
L_{adv}=\mathbb{E}_{(I, F)}[ log(D(y))+log(1-D(G(I)))],
\end{split}
\end{equation}
and with semi-supervised learning, it is defined as
\begin{equation}
\begin{split}
\label{eqn:04b}
L_{adv}=\lambda \mathbb{E}_{(I, F)}[ log(D(y))+log(1-D(G(I)))]+\\
(1-\lambda) \mathbb{E}_{(\hat{I})}( log(1-D(G(\hat{I})))),
\end{split}
\end{equation}
where $\hat{I}$ is an unsupervised data. $G$ is the generator. The output of the discriminator $D$ represents the probability of input image is the real image. $\lambda$ is a weighting hyper-parameter and the expectation value is over a supervised training set $(I, F)$ of input-output pairs. 


\subsection{Implementation Details}
Our proposed ARGAN is implemented in Tensorflow on a computer with Intel(R) Xeon(R) Silver 4114 CPU @ 2.20GHz 192G RAM NVIDIA GeForce GTX 1080Ti. In our experiments, the input size of image is $256\times 256$. We set $N=3$ and $\lambda=0.7$. The minibatch size of 4. The initial learning rate is set as 0.0002. We use Momentum Optimizer to optimize our generator and use Adam Optimizer for the discriminator. We alternatively train the generative network and the discriminative network for 100,000 epochs.

\begin{figure*}[htb]
\vspace{-0.5cm}
 \centering
 \includegraphics[width=0.105\textwidth, height=0.065\textwidth]{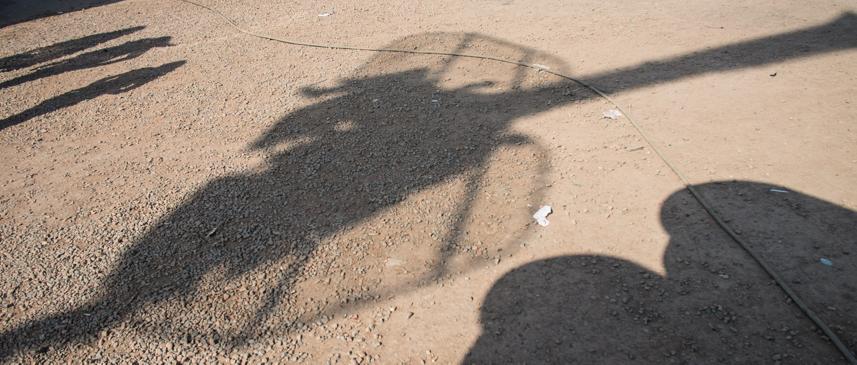}
 \includegraphics[width=0.105\textwidth, height=0.065\textwidth]{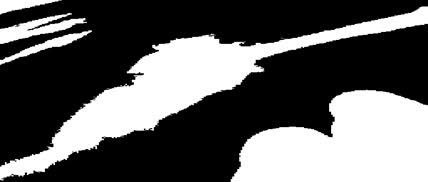}
 \includegraphics[width=0.105\textwidth, height=0.065\textwidth]{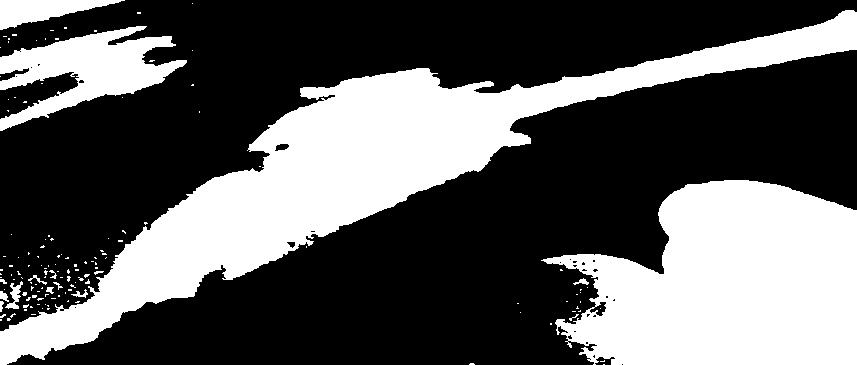}
 \includegraphics[width=0.105\textwidth, height=0.065\textwidth]{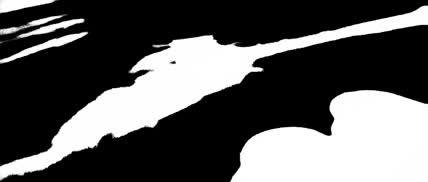}
 \includegraphics[width=0.105\textwidth, height=0.065\textwidth]{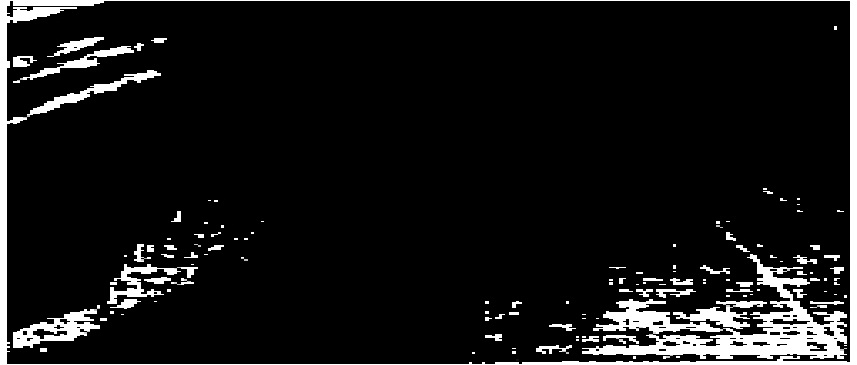}
 \includegraphics[width=0.105\textwidth, height=0.065\textwidth]{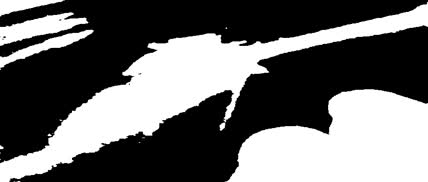}
 \includegraphics[width=0.105\textwidth, height=0.065\textwidth]{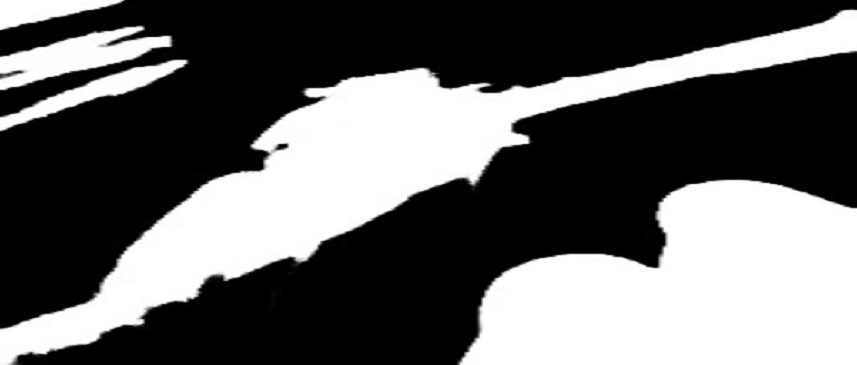}
 \includegraphics[width=0.105\textwidth, height=0.065\textwidth]{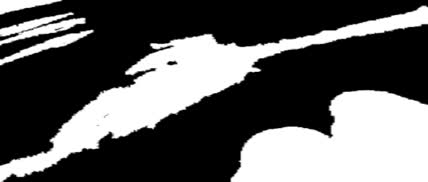}
  \includegraphics[width=0.105\textwidth, height=0.065\textwidth]{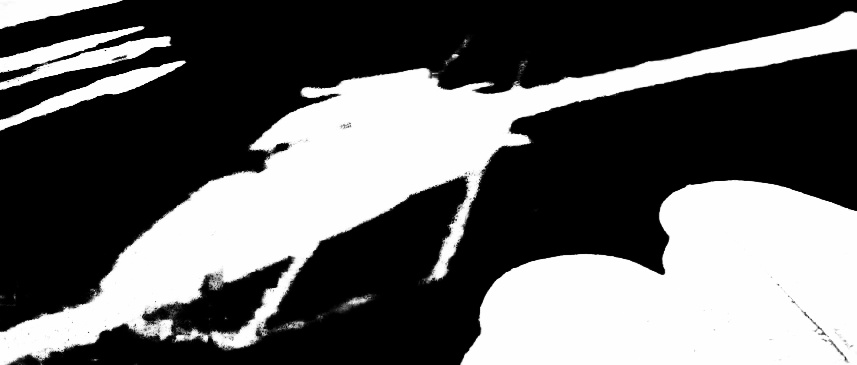}\\
    \subfigure[]{\includegraphics[width=0.105\textwidth]{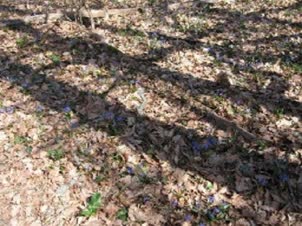}}
 \subfigure[]{\includegraphics[width=0.105\textwidth]{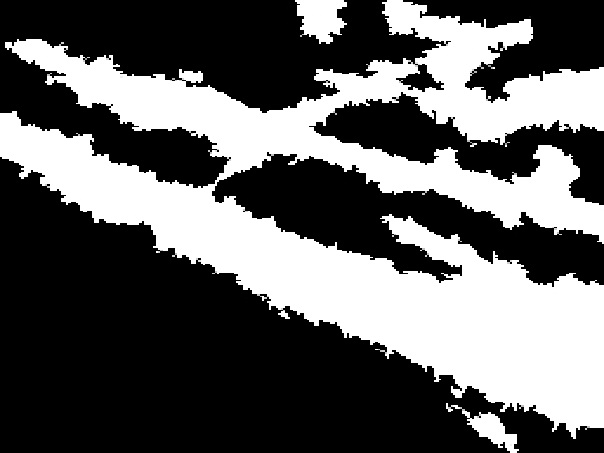}}
  \subfigure[]{\includegraphics[width=0.105\textwidth]{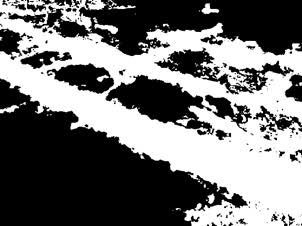}}
  \subfigure[]{\includegraphics[width=0.105\textwidth]{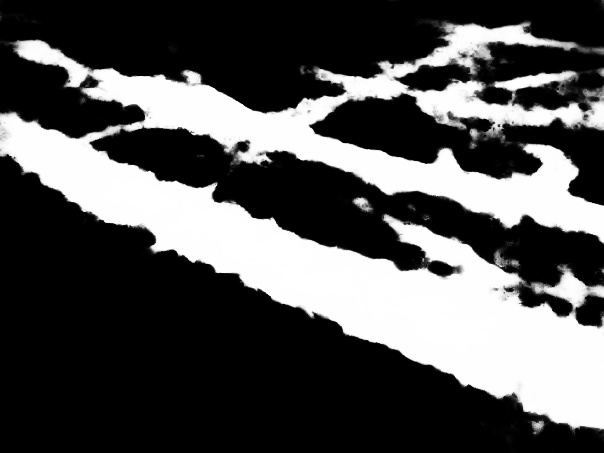}}
  \subfigure[]{\includegraphics[width=0.105\textwidth]{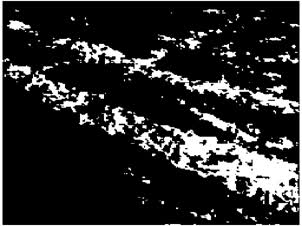}}
  \subfigure[]{\includegraphics[width=0.105\textwidth]{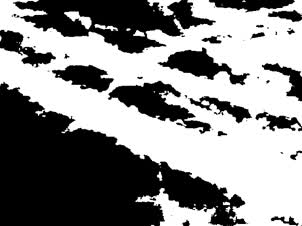}}
 \subfigure[]{\includegraphics[width=0.105\textwidth]{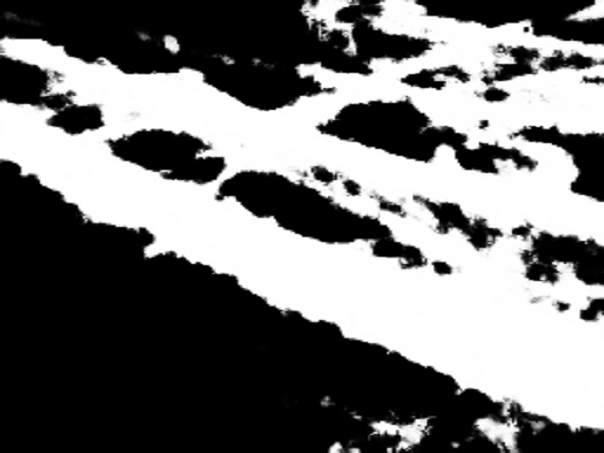}}
 \subfigure[]{\includegraphics[width=0.105\textwidth]{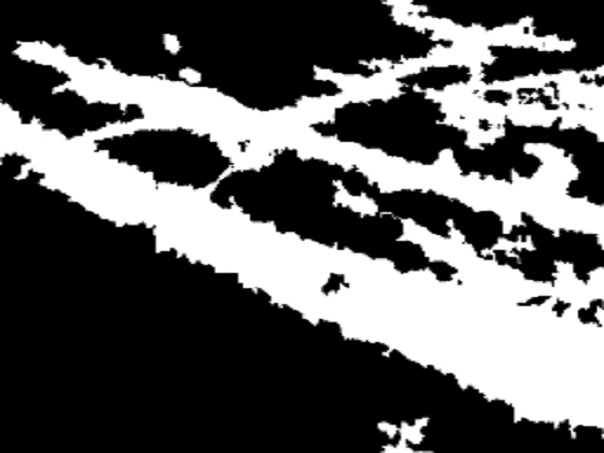}}
   \subfigure[]{\includegraphics[width=0.105\textwidth]{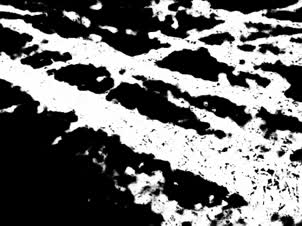}}
    \caption{Shadow detection results comparisons. (a) is input images. (b) is results of Guo~\cite{guo2011single}. (c) is results of Zhang~\cite{zhang2015shadow}. (d) is results of DSC~\cite{hu2018direction}. (e) is results of ST-CGAN~\cite{wang2018stacked}. (f) is results of A+D Net~\cite{le2018a}. (g) is results of BDRAR~\cite{zhu2018bidirectional}. (h) is {{ground-truth}}. (i) is our ARGAN's results.}
    \label{fig5}
    \vspace{-0.2cm}
\end{figure*}

\begin{figure*}[htb]
 \centering
 \includegraphics[width=0.119\textwidth]{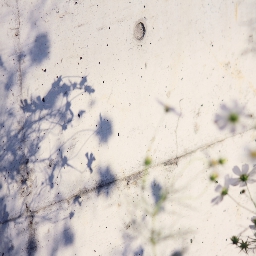}
 \includegraphics[width=0.119\textwidth]{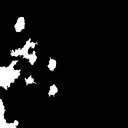}
 \includegraphics[width=0.119\textwidth]{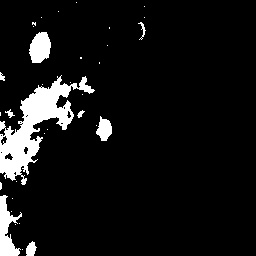}
 \includegraphics[width=0.119\textwidth]{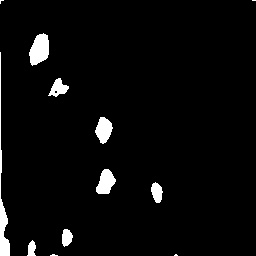}
 \includegraphics[width=0.119\textwidth]{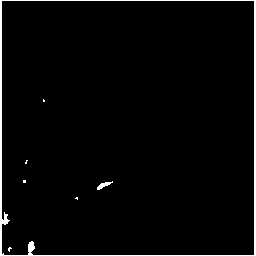}
 \includegraphics[width=0.119\textwidth]{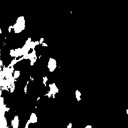}
 \includegraphics[width=0.119\textwidth]{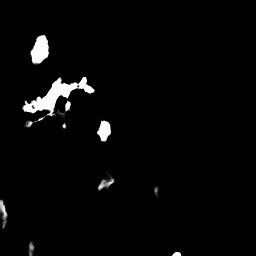}
 \includegraphics[width=0.119\textwidth]{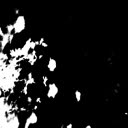}\\
  \subfigure[]{\includegraphics[width=0.119\textwidth]{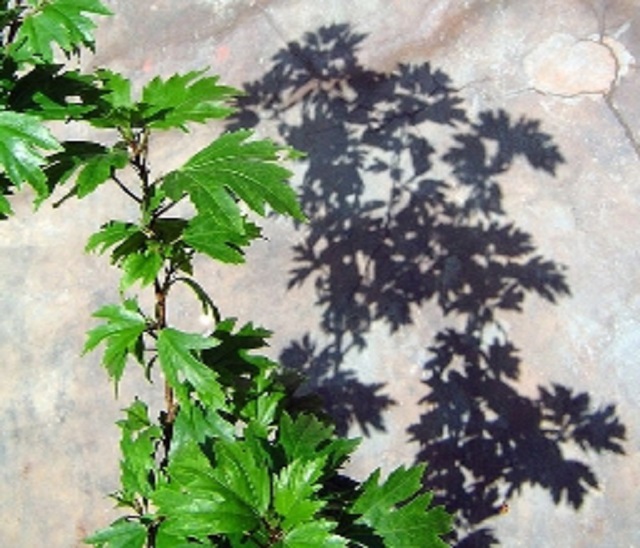}}
 \subfigure[]{\includegraphics[width=0.119\textwidth]{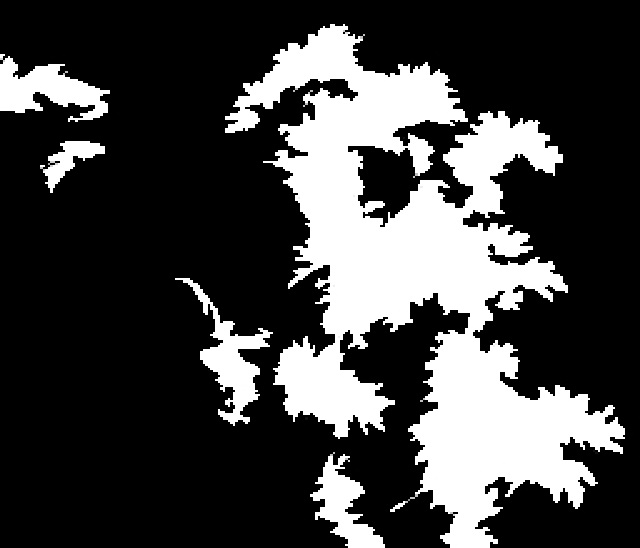}}
 \subfigure[]{\includegraphics[width=0.119\textwidth]{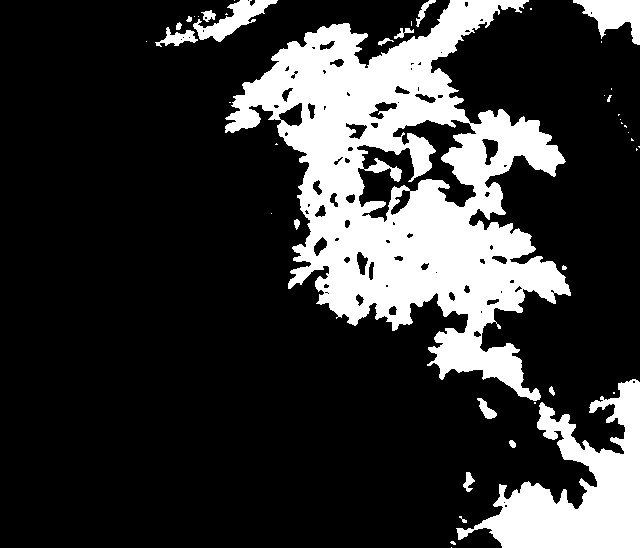}}
 \subfigure[]{\includegraphics[width=0.119\textwidth]{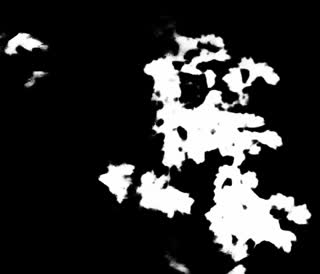}}
 \subfigure[]{\includegraphics[width=0.119\textwidth]{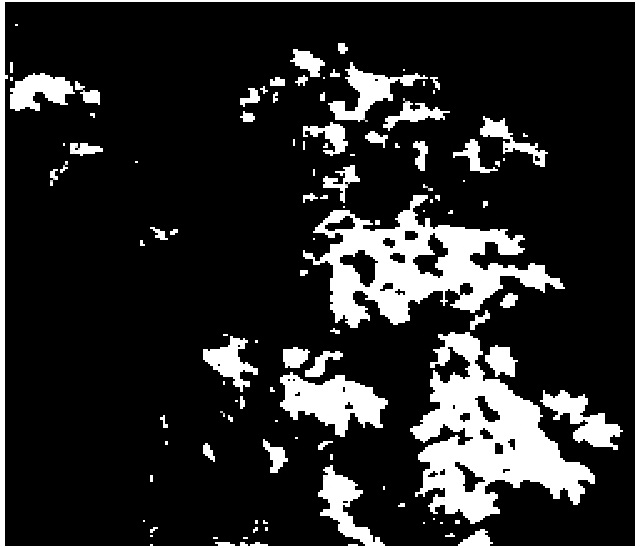}}
 \subfigure[]{\includegraphics[width=0.119\textwidth]{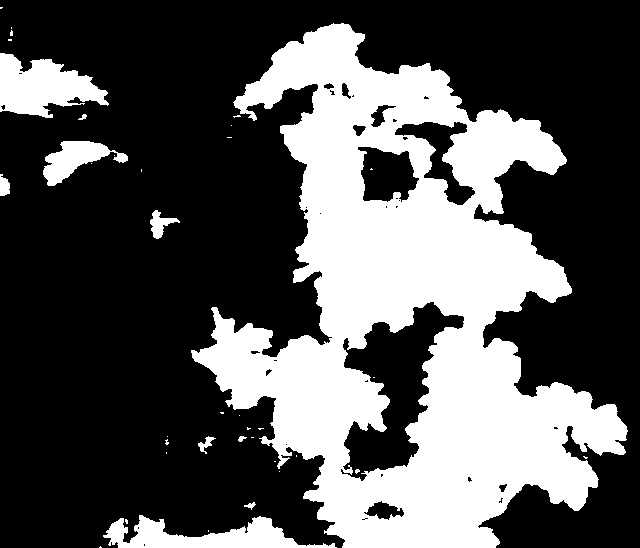}}
 \subfigure[]{\includegraphics[width=0.119\textwidth]{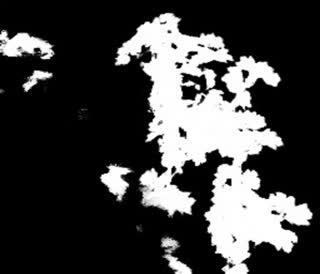}}
 \subfigure[]{\includegraphics[width=0.119\textwidth]{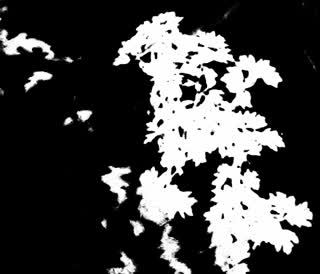}}
    \caption{Shadow detection results comparisons. (a) is input images. (b) is results of Guo~\cite{guo2011single}. (c) is results of Zhang~\cite{zhang2015shadow}. (d) is results of DSC~\cite{hu2018direction}. (e) is results of ST-CGAN~\cite{wang2018stacked}. (f) is results of A+D Net~\cite{le2018a}. (g) is results of BDRAR~\cite{zhu2018bidirectional}. (h) is our ARGAN's results.}
    \label{fig6}
    \vspace{-0.3cm}
\end{figure*}

\section{Experiments}
To verify the effectiveness of our proposed ARGAN+SS, we compare our method with several state-of-the-art shadow detection and removal methods on four datasets: SBU dataset~\cite{Vicente2016Large},  UCF dataset~\cite{zhu2010learning}, SRD dataset~\cite{qu2017deshadownet}, and  ISTD dataset~\cite{wang2018stacked}. 

\subsection{Datasets and metrics}
SBU dataset~\cite{Vicente2016Large} contains 4727 pairs of shadow and shadow mask image. UCF dataset~\cite{zhu2010learning} contains 110 images with corresponding shadow masks. Both datasets have no shadow-free images so that we can only evaluate the detection performance on them. SRD dataset~\cite{qu2017deshadownet} has 408 pairs of shadow and shadow-free images publicly available but, without the ground-truth shadow masks, can only be used for the evaluation of shadow removal. ISTD dataset~\cite{wang2018stacked} contains 1870 image triplets of shadow image, shadow mask, and shadow-free image. We can use this dataset to train our ARGAN and evaluate the performance on both shadow detection and shadow removal. We use 1330 triplets of shadow image, shadow mask and shadow-free images of ISTD~\cite{wang2018stacked} train dataset for training as supervised data, and the remaining 540 triplets for evaluation. 


For shadow detection, we employ Balance Error Rate (BER) \cite{Nguyen2017Shadow} between the ground-truth mask and the predicted shadow matte to evaluate the shadow detection performance. 
For shadow removal, we utilize the root mean square error (RMSE) in “LAB” color space between the recovered shadow removal result and the ground-truth image. 


\subsection{Performance comparison on shadow detection}
We compare our shadow detection results with some state-of-the-art shadow detection methods including two traditional methods,  {\em i.e.}, Guo~\cite{guo2011single} and Zhang~\cite{zhang2015shadow}, and four recent deep learning methods, {\em i.e.}, ST-CGAN~\cite{wang2018stacked}, DSC \cite{hu2018direction}, A+D Net~\cite{le2018a} and BDRAR \cite{zhu2018bidirectional}. To further verify the effectiveness of the LSTM layer in the shadow attention detector, we remove all the LSTM layers and get a variant network which we call ``AGAN". For the fair comparison, we train all the competing models together with our proposed ARGAN on the same training data in supervised learning, and evaluate the shadow detection performance on the SBU, UCF and ISTD datasets. We also collect 1330 images of a wide variety of scenes from online and and take them as unsupervised data to train our model in semi-supervised learning. We denote this method as ``ARGAN+SS" where ``SS" represents semi-supervised.


The results are summarized in Table~\ref{tab1}. As we can observe, (1) among all the competing methods, our ARGAN works the best BER on all the three datasets, which strongly demonstrates that our proposed ARGAN is able to detect accurate shadow regions; (2) without LSTM layers, AGAN performs much worse than ARGAN, which indirectly verify the effectiveness of the LSTM layers in our shadow attention detector; (3) with semi-supervised learning, ARGAN+SS further improve the performance from ARGAN, which strongly proves the robustness of our proposed model.

\begin{figure*}[htb]
\vspace{-0.5cm}
 \centering
 \includegraphics[width=0.119\textwidth]{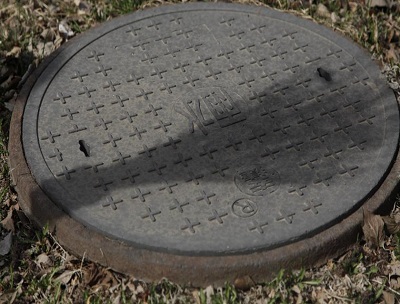}
\includegraphics[width=0.119\textwidth]{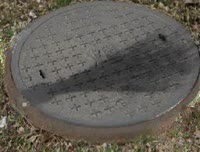}
\includegraphics[width=0.119\textwidth]{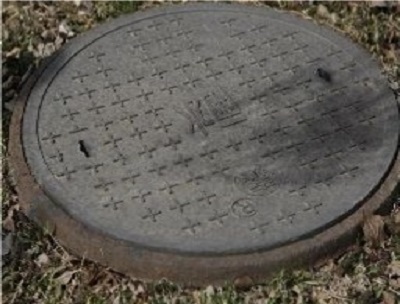}
\includegraphics[width=0.119\textwidth]{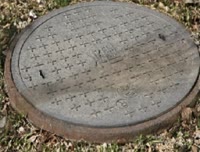}
\includegraphics[width=0.119\textwidth]{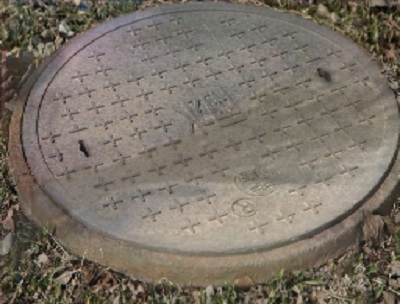}
\includegraphics[width=0.119\textwidth]{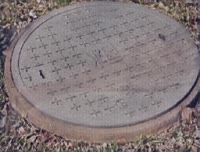}
\includegraphics[width=0.119\textwidth]{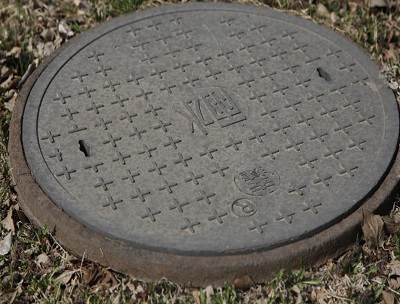}
\includegraphics[width=0.119\textwidth]{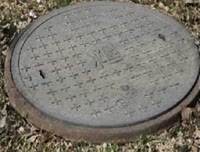}\\
 \subfigure[]{\includegraphics[width=0.119\textwidth]{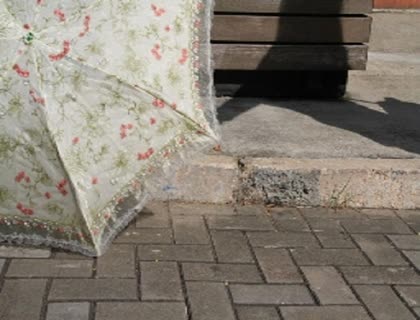}}
  \subfigure[]{\includegraphics[width=0.119\textwidth]{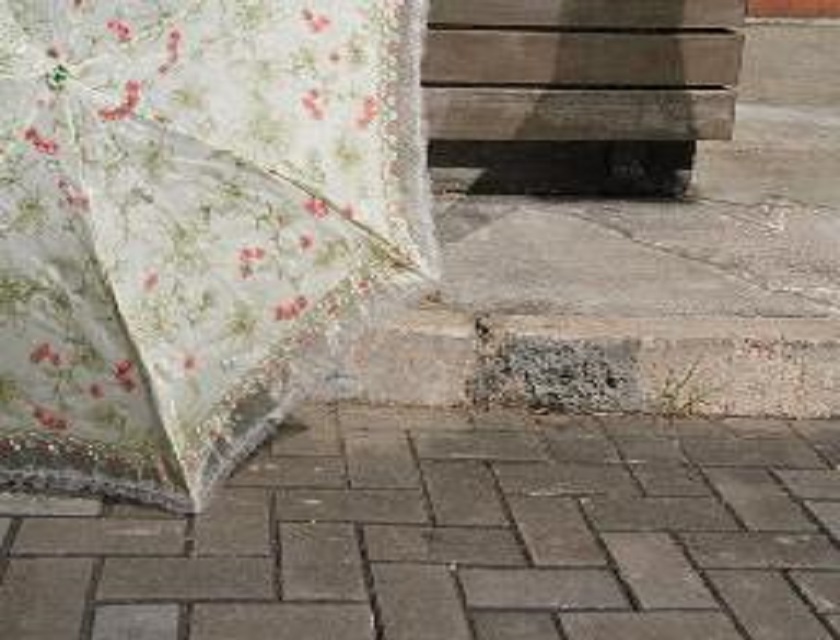}}
 \subfigure[]{\includegraphics[width=0.119\textwidth]{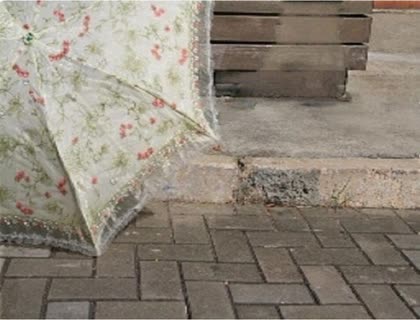}}
 \subfigure[]{\includegraphics[width=0.119\textwidth]{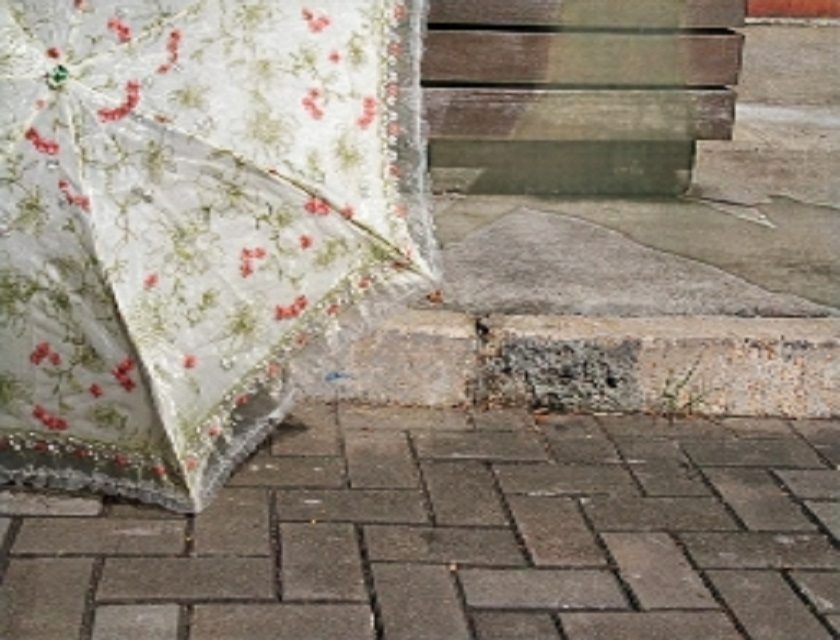}}
   \subfigure[]{\includegraphics[width=0.119\textwidth]{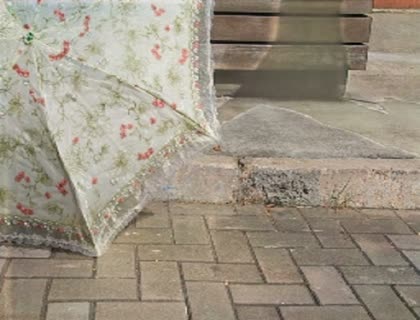}}
  \subfigure[]{\includegraphics[width=0.119\textwidth]{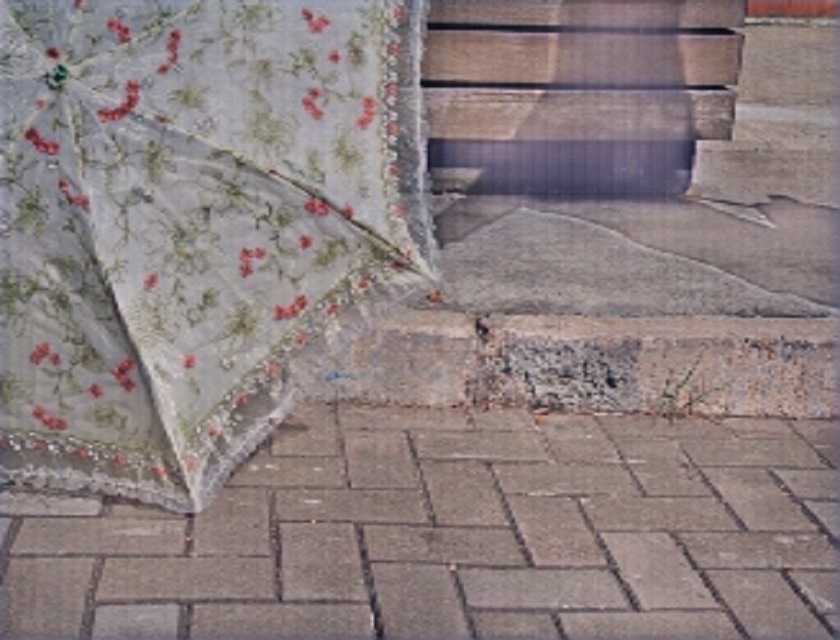}}
  \subfigure[]{\includegraphics[width=0.119\textwidth]{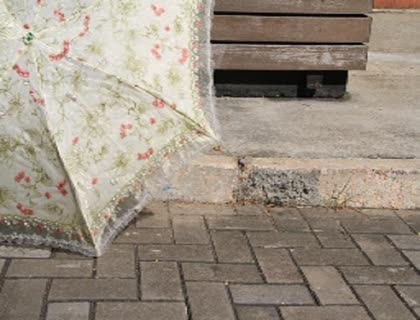}}
   \subfigure[]{\includegraphics[width=0.119\textwidth]{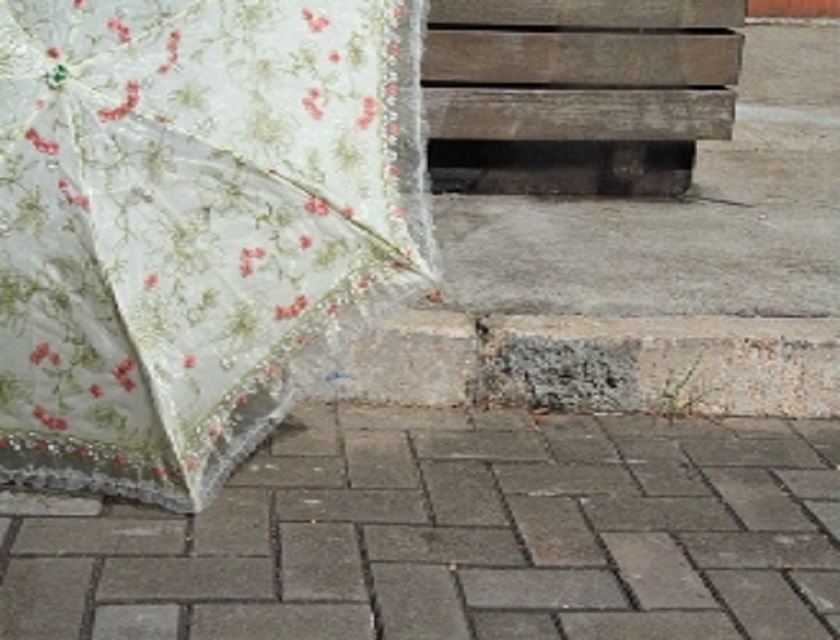}}
%
    \caption{Shadow removal results. (a) is the input images. (b) is results of Guo~\cite{guo2011single}. (c) is the results of Zhang~\cite{zhang2015shadow}. (d) is results of DeshadowNet~\cite{qu2017deshadownet}. (e) is results of DSC~\cite{hu2018direction}. (f) is results of ST-CGAN~\cite{wang2018stacked}. (g) is ground-truth. (h) is our ARGAN's results.}
    \label{fig7}
\vspace{-0.40cm}
\end{figure*}


To further explain the outperforming of our approach, we visualize some results in Figure~\ref{fig5}. 
As we can see, (1) traditional methods Guo~\cite{guo2011single} and Zhang~\cite{zhang2015shadow} are not able to effectively detect slender shadows in the image; 
(2) among all deep learning methods, comparing with ST-CGAN~\cite{wang2018stacked}, DSC~\cite{hu2018direction}, A+D Net~\cite{le2018a} and BDRAR~\cite{zhu2018bidirectional},
our proposed ARGAN is able to detect more accurate shadow regions and even more close to our human observation. 

Figure~\ref{fig6} presents two more shadow images with more complex scenes. Apparently, our ARGAN achieves the best performance on shadow detection. This can be explained by the fact that the shadow attention detectors with recurrent units LSTMs keep updating the detection results gradually from coarse to fine in multiple progressive steps.

\begin{table}[h]
\centering
\footnotesize
\begin{tabular}{c|c|c|c|c}
\hline
\quad  \quad & Year & SBU & UCF &  ISTD\\
\hline
Guo& 2011 &25.03&28.32&27.16\\
Zhang& 2015 &7.13 &9.21&8.56\\
\hline
DSC& 2018 &{5.31}&8.73&2.40\\
{ST-CGAN}& {2018} &{13.56}&{ 17.69}&{3.84}\\
A+D Net& 2018 &7.67&11.05&2.97\\
BDRAR& 2018 &6.61&{9.45} &2.20 \\
\hline
AGAN & 2019 & 7.24&8.67 &4.23\\
ARGAN & 2019 &\textcolor[rgb]{0,0,1}{3.09}&\textcolor[rgb]{0,0,1}{3.76}&\textcolor[rgb]{0,0,1}{2.01}\\
ARGAN+SS & 2019 &\textcolor[rgb]{1,0,0}{2.56}&\textcolor[rgb]{1,0,0}{3.03}&\textcolor[rgb]{1,0,0}{1.75}\\ \hline
\end{tabular}
\vspace{4pt}
\caption{Quantitative comparison results on shadow detection with BER metric. The best and second best results are marked in red and blue colors, respectively.
}.\label{tab1}
\vspace{-0.30cm}
\end{table}

\subsection{Performance comparison on shadow removal}
We compare our proposed ARGAN with the state-of-the-art methods including the traditional methods, {\em i.e.}, Guo~\cite{guo2011single} and Zhang~\cite{zhang2015shadow} and the recent deep learning methods, {\em i.e.}, DeshadowNet~\cite{qu2017deshadownet},  DSC~\cite{hu2018direction}, and ST-CGAN~\cite{wang2018stacked}.  We also compare with two variants of our model, {\em i.e.}, AGAN and ARGAN+SS. We evaluate the performance of shadow removal on SRD dataset and ISTD dataset.

The results are summarized in Table ~\ref{tab2}. As we can see, (1) our proposed ARGAN achieves the best RMSE in both shadow regions and the whole image on the two datasets, which suggests that ARGAN is promising to removal shadows and recover more realistic shadow-free images; (2) without LSTM layers, AGAN cannot recover the shadow-free images qualitatively as well as ARGAN. It can be explained by the fact that LSTM layers affect the detected shadow attention map which affects the quality of final recovered shadow-free images; (3) ARGAN+SS constantly improves the performance for ARGAN no matter in shadow regions, non-shadow regions, or even the whole images. Again, this clearly demonstrates that our proposed ARGAN is good at utilizing sufficient unsupervised shadow images in a semi-supervised learning to improve the quality of the generator and guarantee the performance of both shadow detection and shadow-free image recovery.

We continue to analyze the performance comparison with visualization in Figure~\ref{fig7}. 
In these input images, some areas in non-shadow regions are dark in color. As we can observe, Guo~\cite{guo2011single}, Zhang~\cite{zhang2015shadow},  DeshadowNet~\cite{qu2017deshadownet}, ST-CGAN~\cite{wang2018stacked}, DSC~\cite{hu2018direction} , sometimes will consider the areas with dark color as shadow regions, and recover illumination in these dark areas. 
However, the illumination enhancement changes the color of non-shadow regions, which is not desirable. In contrast, our proposed ARGAN takes full account of color information in the whole image and can produce more natural and realistic shadow removal results.

\begin{figure*}[htb]
 \centering
 \includegraphics[width=0.137\textwidth]{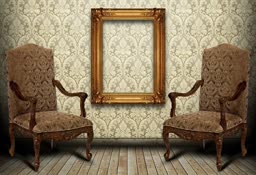}
 \includegraphics[width=0.137\textwidth]{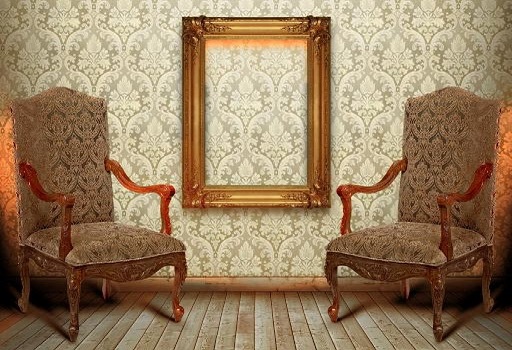}
 \includegraphics[width=0.137\textwidth]{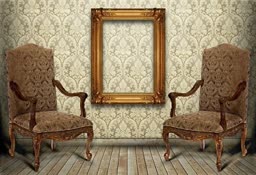}
 \includegraphics[width=0.137\textwidth]{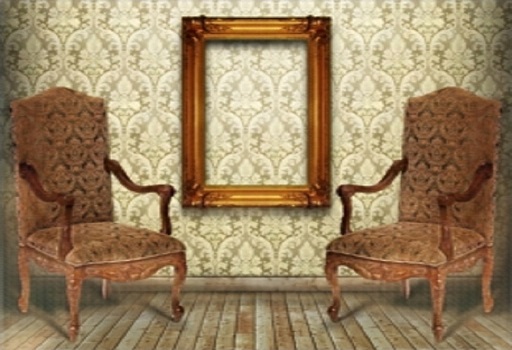}
 \includegraphics[width=0.137\textwidth]{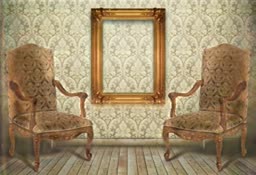}
 \includegraphics[width=0.137\textwidth]{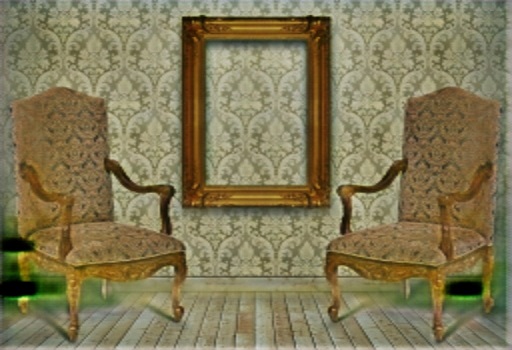}
 \includegraphics[width=0.137\textwidth]{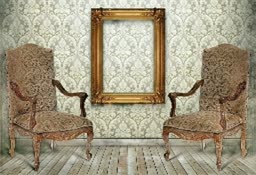}\\
\vspace{4pt}
\includegraphics[width=0.137\textwidth]{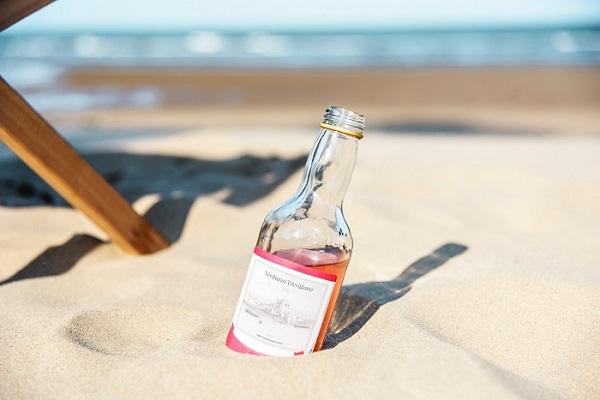}
 \includegraphics[width=0.137\textwidth]{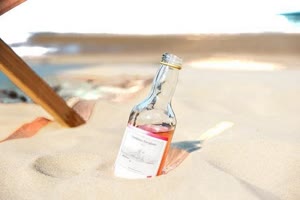}
 \includegraphics[width=0.137\textwidth]{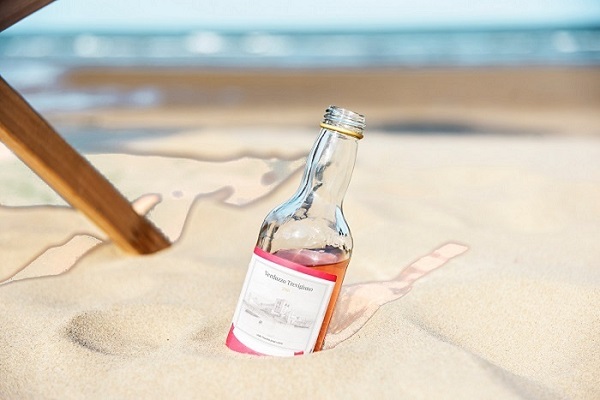}
 \includegraphics[width=0.137\textwidth]{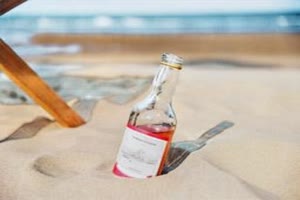}
 \includegraphics[width=0.137\textwidth]{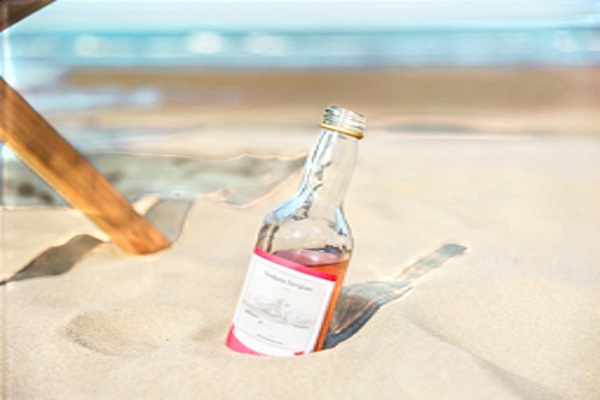}
 \includegraphics[width=0.137\textwidth]{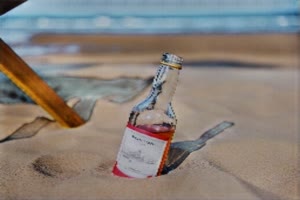}
  \includegraphics[width=0.137\textwidth]{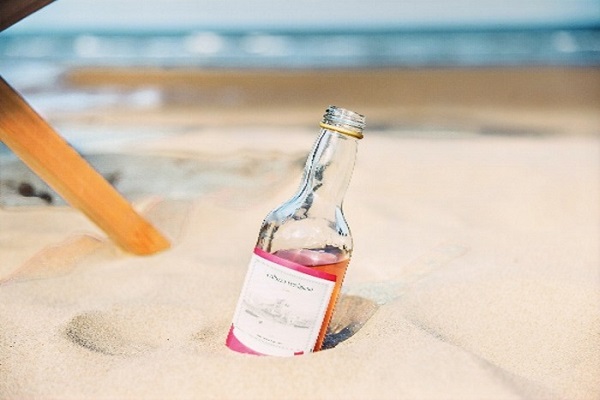}\\
\vspace{4pt}
\includegraphics[width=0.137\textwidth]{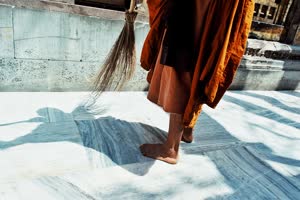}
 \includegraphics[width=0.137\textwidth]{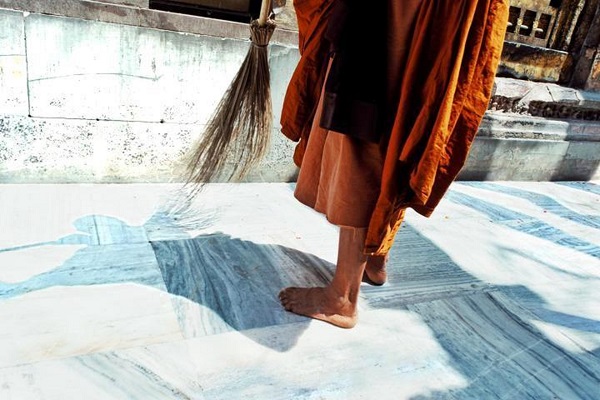}
 \includegraphics[width=0.137\textwidth]{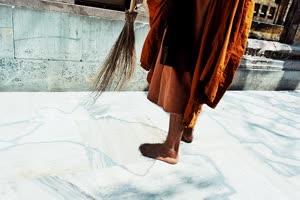}
 \includegraphics[width=0.137\textwidth]{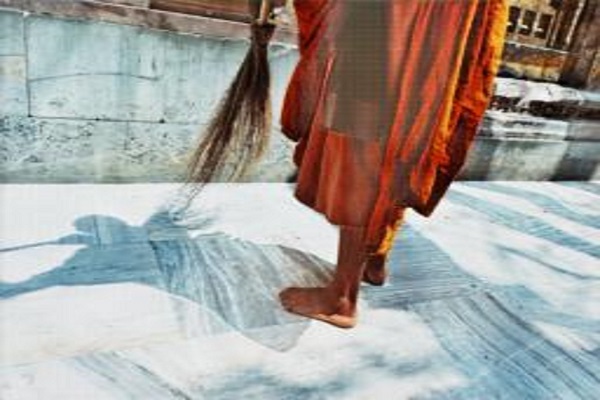}
 \includegraphics[width=0.137\textwidth]{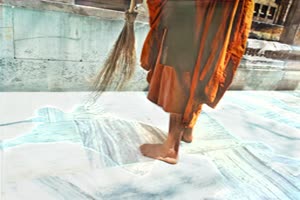}
 \includegraphics[width=0.137\textwidth]{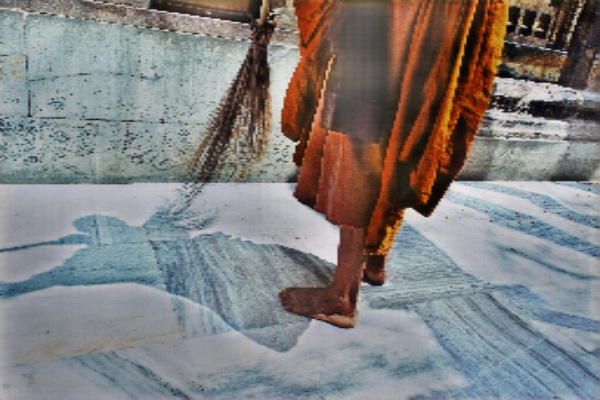}
  \includegraphics[width=0.137\textwidth]{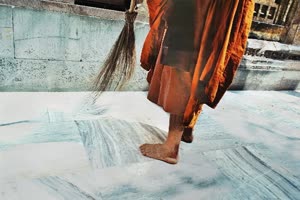}\\

    \subfigure[]{\includegraphics[width=0.137\textwidth]{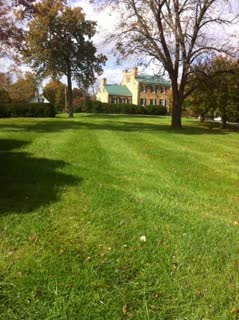}}
  \subfigure[]{\includegraphics[width=0.137\textwidth]{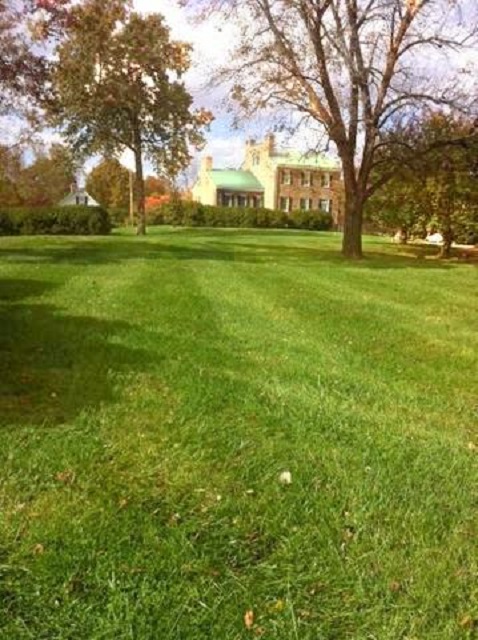}}
 \subfigure[]{\includegraphics[width=0.137\textwidth]{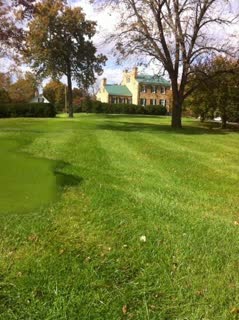}}
 \subfigure[]{\includegraphics[width=0.137\textwidth]{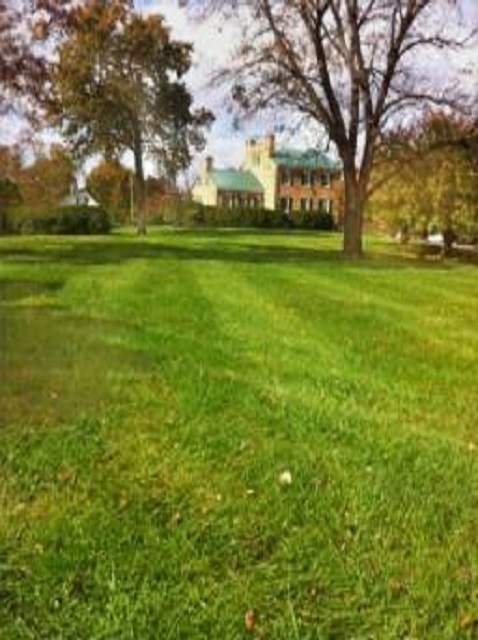}}
   \subfigure[]{\includegraphics[width=0.137\textwidth]{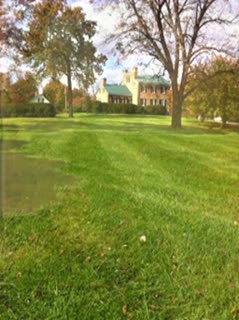}}
  \subfigure[]{\includegraphics[width=0.137\textwidth]{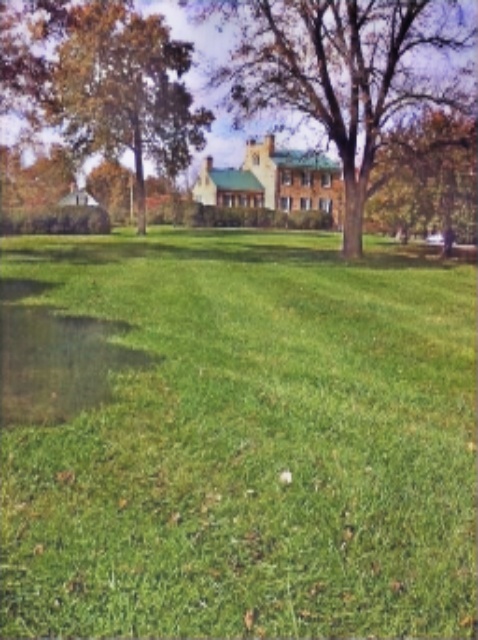}}
  \subfigure[]{\includegraphics[width=0.137\textwidth]{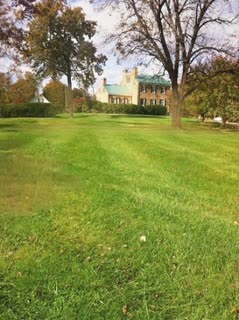}}
    \caption{Shadow removal results. (a) is the input images. (b) is results of Guo~\cite{guo2011single}. (c) is the results of Zhang~\cite{zhang2015shadow}. (d) is results of DeshadowNet~\cite{qu2017deshadownet}. (e) is results of DSC~\cite{hu2018direction}. (f) is results of ST-CGAN~\cite{wang2018stacked}. (g) is our ARGAN's results.}
    \label{fig12}
    \vspace{-0.5cm}
\end{figure*}

 To further verify the robustness of our ARGAN, we choose 7 images with shadows in complex scenes from online and apply all the competing methods to produce shadow-free images, as shown in Figure~\ref{fig12}. Obviously, our ARGAN is robust to deal with shadows with complex scenes. The recovered illumination in original shadow regions in the recovered shadow-free image is consistent with surrounding environment and the texture details in shadow regions are well preserved.

\begin{figure}[htb]
 \centering
  \includegraphics[width=0.09\textwidth]{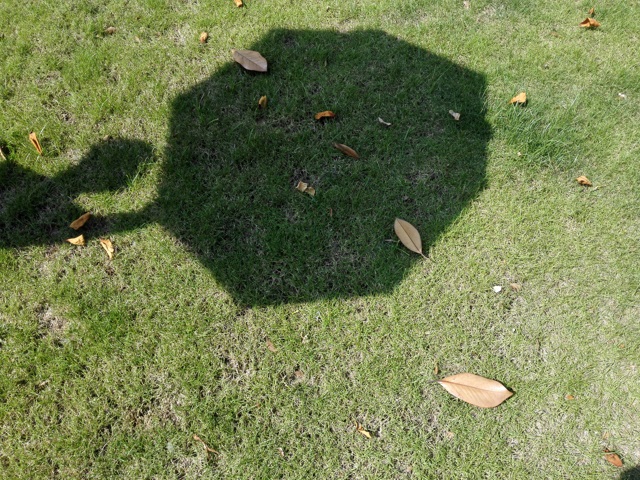}
 \includegraphics[width=0.09\textwidth]{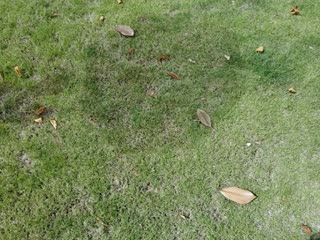}
 \includegraphics[width=0.09\textwidth]{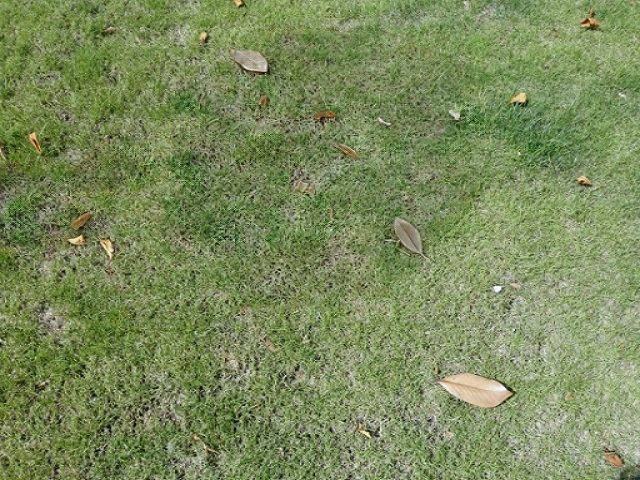}
 \includegraphics[width=0.09\textwidth]{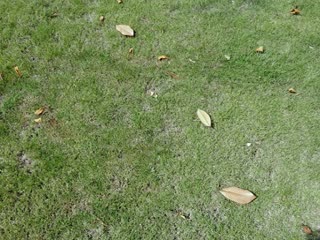}
 \includegraphics[width=0.09\textwidth]{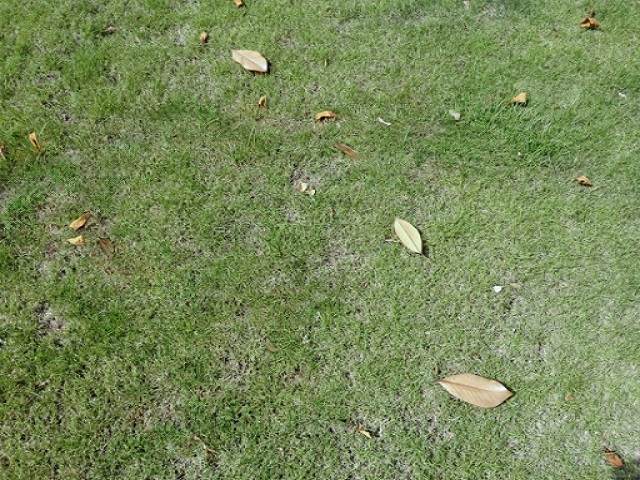}\\
  \subfigure[]{\includegraphics[width=0.09\textwidth]{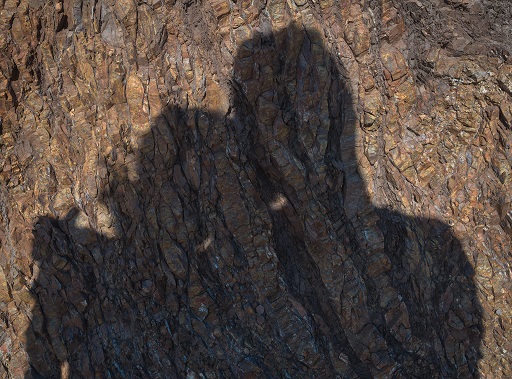}}
 \subfigure[]{\includegraphics[width=0.09\textwidth]{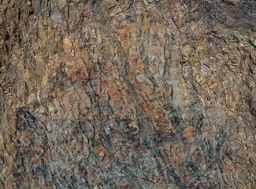}}
 \subfigure[]{\includegraphics[width=0.09\textwidth]{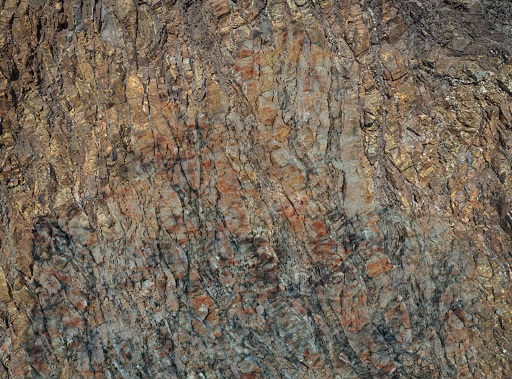}}
 \subfigure[]{\includegraphics[width=0.09\textwidth]{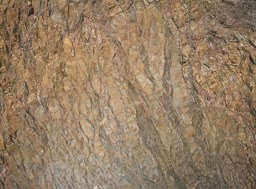}}
 \subfigure[]{\includegraphics[width=0.09\textwidth]{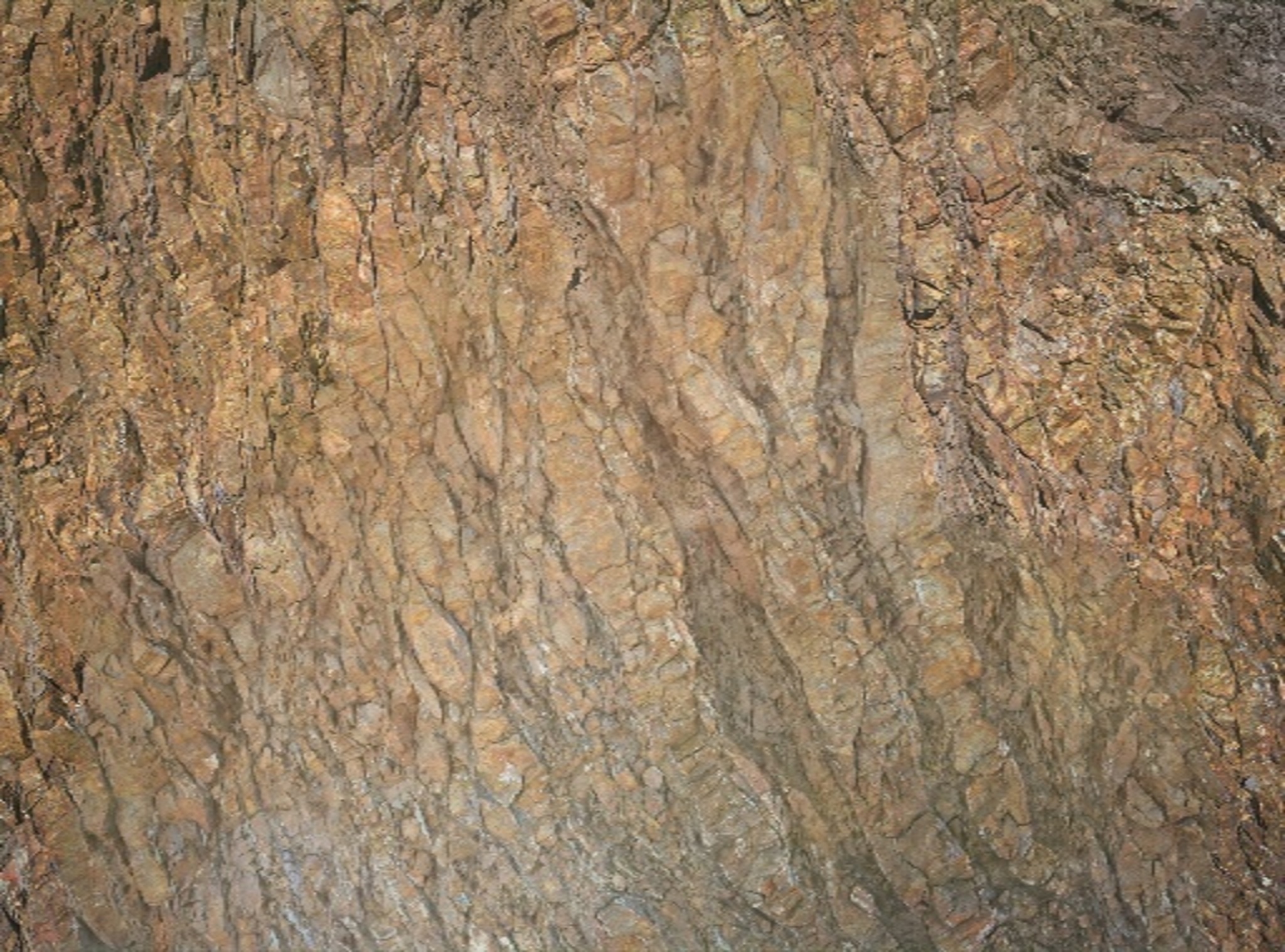}}
    \caption{The shadow removal results of recurrent N in a different value. (a) is the input image. (b), (c), (d) and (e) are the results of N taking 1, 2, 3, 4 respectively.}
    \label{fig17}
    \vspace{-0.3cm}
\end{figure}

\begin{figure}[htb]
 \centering
  \includegraphics[width=0.115\textwidth]{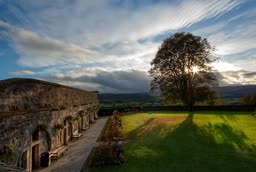}
 \includegraphics[width=0.115\textwidth]{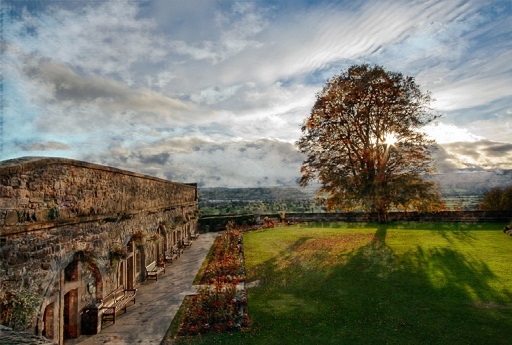}
 \includegraphics[width=0.115\textwidth]{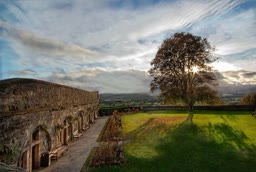}
 \includegraphics[width=0.115\textwidth]{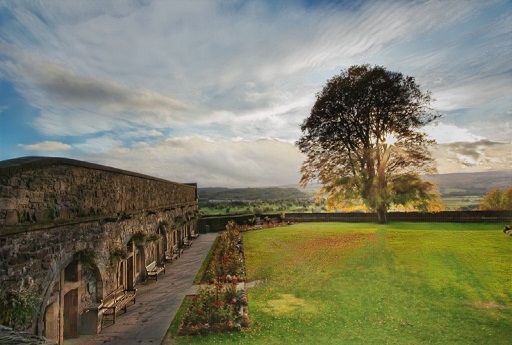}\\
    \caption{The visualization results of ablation analysis. (a) is the input images. (b) is shadow removal results without using LSTM. (c) is shadow removal results without using unsupervised data. (d) is our final shadow removal results.}
    \label{fig18}
    \vspace{-0.3cm}
\end{figure}

\begin{table}[h]
\centering
\footnotesize
\begin{tabular}{c|ccc|ccc}
\hline
 & \multicolumn{3}{c}{SRD}&  \multicolumn{3}{c}{ISTD} \\ \hline
&S&N&A&S&N&A \\
\hline
Guo
&29.89&{6.47}&12.60&18.95&{7.46}&9.3 \\
Zhang
&{9.56}&6.9&7.24&{9.77}&7.12&8.16 \\
\hline
DeshadowNet
& 17.96 & 6.53 & 8.47 & 12.76 & {7.19} & {7.83} \\
ST-CGAN
& 18.64 & 6.37 & 8.23 & 10.31 & {6.92} & {7.46} \\
DSC
&11.31&{6.72}&{7.83}&9.22&6.50&7.10 \\
\hline
AGAN&14.68&5.94&7.65&11.79&6.33&7.57 \\
ARGAN&\textcolor[rgb]{0,0,1}{7.24}&\textcolor[rgb]{0,0,1}{4.71}&\textcolor[rgb]{0,0,1}{5.74}&\textcolor[rgb]{0,0,1}{7.21}&\textcolor[rgb]{0,0,1}{5.83}&\textcolor[rgb]{0,0,1}{6.68} \\
ARGAN+SS&\textcolor[rgb]{1,0,0}{6.35}&\textcolor[rgb]{1,0,0}{4.46}&\textcolor[rgb]{1,0,0}{5.31}&\textcolor[rgb]{1,0,0}{6.65}&\textcolor[rgb]{1,0,0}{5.41}&\textcolor[rgb]{1,0,0}{5.89} \\\hline
\end{tabular}
\vspace{4pt}
\caption{Quantitative comparison results on shadow removal with RMSE metric. The best and second best results are marked in red and blue colors, respectively. In the table, S represents shadow regions, N represents non-shadow regions, and A represents the whole image. 
} \label{tab2}
\vspace{-0.30cm}
\end{table}

\subsection{Discussion}
To further explore how the value of $N$ affects the final performance, we take experiments $N=1,2,3$ and $4$ to generate a shadow-free images by our ARGAN. On ISTD, the BER values for shadow detection are 2.22, 2.08, 2.01 and 2.01, respectively; and the RMSE values for shadow removal on the whole images are 7.35, 6.97, 6.68 and 6.67, respectively. We observe that $N=3$ is a good trade-off between performance and complexity, as shown in Figure~\ref{fig17}.


In addition, we also visualize the shadow-removal results with AGAN, ARGAN and ARGAN+SS in Figure~\ref{fig18}. We can observe that LSTM layers really matter for shadow-free image recovery and with semi-supervised strategy and that ARGAN+SS is powerful to handle shadow images with complex scenes.



\section{Conclusions}
In this paper, we have proposed a robust attentive recurrent generative adversarial network for shadow detection and removal. The generator both to generate shadow attention maps and to recover the shadow-removal images involves multiple progressive steps in a coarse-to-fine fashion. Our model is able to handle shadows with complex scenes and is very flexible to incorporate sufficient unsupervised shadow images to train a powerful model.
The future work includes extending our method to video shadow detection and removal.

\section*{Acknowledgement}
This work was partly supported by The National Key Research and Development Program of China (2017YFB1002600), the NSFC (No. 61672390), and Key Technological Innovation Projects of Hubei Province (2018AAA062). The corresponding author is Chunxia Xiao.

{\small
\bibliographystyle{ieee}
\bibliography{SARGAN_CameraReady}

\begin{thebibliography}{10}\itemsep=-1pt

\bibitem{anderson2018bottom}
P.~Anderson, X.~He, C.~Buehler, D.~Teney, M.~Johnson, S.~Gould, and L.~Zhang.
\newblock Bottom-up and top-down attention for image captioning and visual
  question answering.
\newblock In {\em IEEE Conference on Computer Vision and Pattern Recognition
  (CVPR)}, volume~3, page~6, 2018.

\bibitem{bahdanau2014neural}
D.~Bahdanau, K.~Cho, and Y.~Bengio.
\newblock Neural machine translation by jointly learning to align and
  translate.
\newblock {\em arXiv preprint arXiv:1409.0473}, 2014.

\bibitem{chorowski2015attention}
J.~K. Chorowski, D.~Bahdanau, D.~Serdyuk, K.~Cho, and Y.~Bengio.
\newblock Attention-based models for speech recognition.
\newblock In {\em Advances in neural information processing systems (NeurIPS)},
  2015.

\bibitem{chuang2003shadow}
Y.-Y. Chuang, D.~B. Goldman, B.~Curless, D.~H. Salesin, and R.~Szeliski.
\newblock Shadow matting and compositing.
\newblock In {\em ACM Transactions on Graphics (TOG)}, volume~22, pages
  494--500, 2003.

\bibitem{Cucchiara2002Improving}
R.~Cucchiara, C.~Grana, M.~Piccardi, A.~Prati, and S.~Sirotti.
\newblock Improving shadow suppression in moving object detection with hsv
  color information.
\newblock In {\em Intelligent Transportation Systems}, 2002.

\bibitem{Eli2011Shadow}
A.~Eli and H.~O. Hagit.
\newblock Shadow removal using intensity surfaces and texture anchor points.
\newblock {\em IEEE Transactions on Pattern Analysis Machine Intelligence
  (T-PAMI)}, 33(6):1202--1216, 2011.

\bibitem{Feng2008Texture}
L.~Feng and M.~Gleicher.
\newblock Texture-consistent shadow removal.
\newblock In {\em European Conference on Computer Vision (ECCV)}, 2008.

\bibitem{fu2017removing}
X.~Fu, J.~Huang, D.~Zeng, Y.~Huang, X.~Ding, and J.~Paisley.
\newblock Removing rain from single images via a deep detail network.
\newblock In {\em IEEE Conference on Computer Vision and Pattern Recognition
  (CVPR)}, 2017.

\bibitem{gao2019hierarchical}
L.~Gao, X.~Li, J.~Song, and H.~T. Shen.
\newblock Hierarchical lstms with adaptive attention for visual captioning.
\newblock {\em IEEE Transactions on Pattern Analysis Machine Intelligence
  (T-PAMI)}, 2019.

\bibitem{goodfellow2014generative}
I.~Goodfellow, J.~Pouget-Abadie, M.~Mirza, B.~Xu, D.~Warde-Farley, S.~Ozair,
  A.~Courville, and Y.~Bengio.
\newblock Generative adversarial nets.
\newblock In {\em Advances in Neural Information Processing Systems (NeurIPS)},
  2014.

\bibitem{Gryka2015Learning}
M.~Gryka, M.~Terry, and G.~J. Brostow.
\newblock Learning to remove soft shadows.
\newblock {\em ACM Transactions on Graphics (TOG)}, 34(5):1--15, 2015.

\bibitem{guo2011single}
R.~Guo, Q.~Dai, and D.~Hoiem.
\newblock Single-image shadow detection and removal using paired regions.
\newblock In {\em IEEE Conference on Computer Vision and Pattern Recognition
  (CVPR)}, 2011.

\bibitem{Han2016StackGAN}
Z.~Han, X.~Tao, and H.~Li.
\newblock Stackgan: Text to photo-realistic image synthesis with stacked
  generative adversarial networks.
\newblock 2016.

\bibitem{he2016deep}
K.~He, X.~Zhang, S.~Ren, and J.~Sun.
\newblock Deep residual learning for image recognition.
\newblock In {\em IEEE Conference on Computer Vision and Pattern Recognition
  (CVPR)}, 2016.

\bibitem{hochreiter1997long}
S.~Hochreiter and J.~Schmidhuber.
\newblock Long short-term memory.
\newblock {\em Neural Computation}, 9(8):1735--1780, 1997.

\bibitem{hu2018direction}
X.~Hu, C.-W. Fu, L.~Zhu, J.~Qin, and P.-A. Heng.
\newblock Direction-aware spatial context features for shadow detection and
  removal.
\newblock {\em arXiv preprint arXiv}, 2018.

\bibitem{Hu2017Direction}
X.~Hu, L.~Zhu, C.-W. Fu, J.~Qin, and P.-A. Heng.
\newblock Direction-aware spatial context features for shadow detection.
\newblock 2018.

\bibitem{hua2013collaborative}
G.~Hua, C.~Long, M.~Yang, and Y.~Gao.
\newblock Collaborative active learning of a kernel machine ensemble for
  recognition.
\newblock In {\em IEEE International Conference on Computer Vision (ICCV)},
  2013.

\bibitem{hua2018collaborative}
G.~Hua, C.~Long, M.~Yang, and Y.~Gao.
\newblock Collaborative active visual recognition from crowds: A distributed
  ensemble approach.
\newblock {\em IEEE transactions on pattern analysis and machine intelligence},
  40(3):582--594, 2018.

\bibitem{isola2017image}
P.~Isola, J.-Y. Zhu, T.~Zhou, and A.~A. Efros.
\newblock Image-to-image translation with conditional adversarial networks.
\newblock {\em arXiv preprint}, 2017.

\bibitem{johnson2016perceptual}
J.~Johnson, A.~Alahi, and L.~Fei-Fei.
\newblock Perceptual losses for real-time style transfer and super-resolution.
\newblock In {\em European Conference on Computer Vision (ECCV)}, 2016.

\bibitem{junejo2008estimating}
I.~N. Junejo and H.~Foroosh.
\newblock Estimating geo-temporal location of stationary cameras using shadow
  trajectories.
\newblock In {\em Proceedings of the European Conference on Computer Vision
  (ECCV)}. Springer, 2008.

\bibitem{Karsch2014Automatic}
K.~Karsch, K.~Sunkavalli, S.~Hadap, N.~Carr, H.~Jin, R.~Fonte, M.~Sittig, and
  D.~Forsyth.
\newblock Automatic scene inference for 3d object compositing.
\newblock {\em ACM Transactions on Graphics (TOG)}, 33(3):1--15, 2014.

\bibitem{khan2016automatic}
S.~H. Khan, M.~Bennamoun, F.~Sohel, and R.~Togneri.
\newblock Automatic shadow detection and removal from a single image.
\newblock {\em IEEE Transactions on Pattern Analysis Machine Intelligence
  (T-PAMI)}, (3):431--446, 2016.

\bibitem{kiela2018dynamic}
D.~Kiela, C.~Wang, and K.~Cho.
\newblock Dynamic meta-embeddings for improved sentence representations.
\newblock In {\em Conference on Empirical Methods in Natural Language
  Processing}, 2018.

\bibitem{Lalonde2010Detecting}
J.~F. Lalonde, A.~A. Efros, and S.~G. Narasimhan.
\newblock Detecting ground shadows in outdoor consumer photographs.
\newblock In {\em European Conference on Computer Vision (ECCV)}, 2010.

\bibitem{le2018a}
H.~Le, Y.~Vicente, F.~Tomas, V.~Nguyen, M.~Hoai, and D.~Samaras.
\newblock A+ d net: Training a shadow detector with adversarial shadow
  attenuation.
\newblock In {\em European Conference on Computer Vision (ECCV)}, 2018.

\bibitem{ledig2017photo}
C.~Ledig, L.~Theis, F.~Husz{\'a}r, J.~Caballero, A.~Cunningham, A.~Acosta,
  A.~Aitken, A.~Tejani, J.~Totz, Z.~Wang, et~al.
\newblock Photo-realistic single image super-resolution using a generative
  adversarial network.
\newblock {\em arXiv preprint}, 2017.

\bibitem{li2016precomputed}
C.~Li and M.~Wand.
\newblock Precomputed real-time texture synthesis with markovian generative
  adversarial networks.
\newblock In {\em European Conference on Computer Vision (ECCV)}, 2016.

\bibitem{Li2018Learning}
Z.~Li and N.~Snavely.
\newblock Learning intrinsic image decomposition from watching the world.
\newblock 2018.

\bibitem{liu2017unsupervised}
M.-Y. Liu, T.~Breuel, and J.~Kautz.
\newblock Unsupervised image-to-image translation networks.
\newblock In {\em Advances in Neural Information Processing Systems (NeurIPS)},
  2017.

\bibitem{long2015multi}
C.~Long and G.~Hua.
\newblock Multi-class multi-annotator active learning with robust gaussian
  process for visual recognition.
\newblock In {\em IEEE International Conference on Computer Vision (ICCV)},
  2015.

\bibitem{Long_2017_CVPR}
C.~Long and G.~Hua.
\newblock Correlational gaussian processes for cross-domain visual recognition.
\newblock In {\em IEEE Conference on Computer Vision and Pattern Recognition
  (CVPR)}, 2017.

\bibitem{long2014accurate}
C.~Long, X.~Wang, G.~Hua, M.~Yang, and Y.~Lin.
\newblock Accurate object detection with location relaxation and regionlets
  re-localization.
\newblock In {\em Asian Conference on Computer Vision (ACCV)}, 2014.

\bibitem{lu2017knowing}
J.~Lu, C.~Xiong, D.~Parikh, and R.~Socher.
\newblock Knowing when to look: Adaptive attention via a visual sentinel for
  image captioning.
\newblock In {\em IEEE Conference on Computer Vision and Pattern Recognition
  (CVPR)}, volume~6, page~2, 2017.

\bibitem{lu2016hierarchical}
J.~Lu, J.~Yang, D.~Batra, and D.~Parikh.
\newblock Hierarchical question-image co-attention for visual question
  answering.
\newblock In {\em Advances In Neural Information Processing Systems (NeurIPS)},
  2016.

\bibitem{luo2019end}
W.~Luo, P.~Sun, F.~Zhong, W.~Liu, T.~Zhang, and Y.~Wang.
\newblock End-to-end active object tracking and its real-world deployment via
  reinforcement learning.
\newblock {\em IEEE Transactions on Pattern Analysis Machine Intelligence
  (T-PAMI)}, 2019.

\bibitem{Miki2000Moving}
I.~Miki, P.~C. Cosman, G.~T. Kogut, and M.~M. Trivedi.
\newblock Moving shadow and object detection in traffic scenes.
\newblock In {\em International Conference on Pattern Recognition (ICPR)},
  2000.

\bibitem{miyato2018spectral}
T.~Miyato, T.~Kataoka, M.~Koyama, and Y.~Yoshida.
\newblock Spectral normalization for generative adversarial networks.
\newblock {\em arXiv preprint arXiv}, 2018.

\bibitem{Mohan2007Editing}
A.~Mohan, J.~Tumblin, and P.~Choudhury.
\newblock Editing soft shadows in a digital photograph.
\newblock {\em IEEE Comput Graph Appl}, 27(2):23--31, 2007.

\bibitem{Nguyen2017Shadow}
V.~Nguyen, T.~F.~Y. Vicente, M.~Zhao, M.~Hoai, and D.~Samaras.
\newblock Shadow detection with conditional generative adversarial networks.
\newblock In {\em IEEE International Conference on Computer Vision (ICCV)},
  2017.

\bibitem{Okabe2009Attached}
T.~Okabe, I.~Sato, and Y.~Sato.
\newblock Attached shadow coding: Estimating surface normals from shadows under
  unknown reflectance and lighting conditions.
\newblock In {\em IEEE International Conference on Computer Vision (ICCV)},
  2009.

\bibitem{pathak2016context}
D.~Pathak, P.~Krahenbuhl, J.~Donahue, T.~Darrell, and A.~A. Efros.
\newblock Context encoders: Feature learning by inpainting.
\newblock In {\em IEEE Conference on Computer Vision and Pattern Recognition
  (CVPR)}, 2016.

\bibitem{qian2018attentive}
R.~Qian, R.~T. Tan, W.~Yang, J.~Su, and J.~Liu.
\newblock Attentive generative adversarial network for raindrop removal from a
  single image.
\newblock In {\em IEEE Conference on Computer Vision and Pattern Recognition
  (CVPR)}, 2018.

\bibitem{qu2017deshadownet}
L.~Qu, J.~Tian, S.~He, Y.~Tang, and R.~W. Lau.
\newblock Deshadownet: A multi-context embedding deep network for shadow
  removal.
\newblock In {\em IEEE Conference on Computer Vision and Pattern Recognition
  (CVPR)}, 2017.

\bibitem{reinhard2001color}
E.~Reinhard, M.~Adhikhmin, B.~Gooch, and P.~Shirley.
\newblock Color transfer between images.
\newblock {\em IEEE Computer graphics and applications}, 21(5):34--41, 2001.

\bibitem{shor2008shadow}
Y.~Shor and D.~Lischinski.
\newblock The shadow meets the mask: Pyramid-based shadow removal.
\newblock In {\em Computer Graphics Forum}, volume~27, pages 577--586, 2008.

\bibitem{simo2018mastering}
E.~Simo-Serra, S.~Iizuka, and H.~Ishikawa.
\newblock Mastering sketching: adversarial augmentation for structured
  prediction.
\newblock {\em ACM Transactions on Graphics (TOG)}, 37(1):11, 2018.

\bibitem{simonyan2014very}
K.~Simonyan and A.~Zisserman.
\newblock Very deep convolutional networks for large-scale image recognition.
\newblock {\em arXiv preprint arXiv:1409.1556}, 2014.

\bibitem{tao2019radical}
H.~Tao, S.~Tong, H.~Zhao, T.~Xu, B.~Jin, and Q.~Liu.
\newblock A radical-aware attention-based model for chinese text
  classification.
\newblock In {\em Association for the Advancement of Artificial Intelligence
  (AAAI)}, 2019.

\bibitem{tran2019disentangling}
L.~Tran, J.~Kossaifi, Y.~Panagakis, and M.~Pantic.
\newblock Disentangling geometry and appearance with regularised geometry-aware
  generative adversarial networks.
\newblock {\em International Journal of Computer Vision}, pages 1--21, 2019.

\bibitem{vicente2018leave}
T.~F.~Y. Vicente, M.~Hoai, and D.~Samaras.
\newblock Leave-one-out kernel optimization for shadow detection and removal.
\newblock {\em IEEE Transactions on Pattern Analysis Machine Intelligence
  (T-PAMI)}, 40(3):682--695, 2018.

\bibitem{Vicente2016Large}
T.~F.~Y. Vicente, L.~Hou, C.-P. Yu, M.~Hoai, and D.~Samaras.
\newblock Large-scale training of shadow detectors with noisily-annotated
  shadow examples.
\newblock In {\em European Conference on Computer Vision (ECCV)}, 2016.

\bibitem{wang2018stacked}
J.~Wang, X.~Li, and J.~Yang.
\newblock Stacked conditional generative adversarial networks for jointly
  learning shadow detection and shadow removal.
\newblock In {\em IEEE Conference on Computer Vision and Pattern Recognition
  (CVPR)}, 2018.

\bibitem{wu2007natural}
T.-P. Wu, C.-K. Tang, M.~S. Brown, and H.-Y. Shum.
\newblock Natural shadow matting.
\newblock {\em ACM Transactions on Graphics (TOG)}, 26(2):8, 2007.

\bibitem{xiao2013fast}
C.~Xiao, R.~She, D.~Xiao, and K.-L. Ma.
\newblock Fast shadow removal using adaptive multi-scale illumination transfer.
\newblock In {\em Computer Graphics Forum}, volume~32, pages 207--218, 2013.

\bibitem{xiao2014shadow}
Y.~Xiao, E.~Tsougenis, and C.-K. Tang.
\newblock Shadow removal from single rgb-d images.
\newblock In {\em IEEE Conference on Computer Vision and Pattern Recognition
  (CVPR)}, 2014.

\bibitem{xingjian2015convolutional}
S.~Xingjian, Z.~Chen, H.~Wang, D.-Y. Yeung, W.-K. Wong, and W.-c. Woo.
\newblock Convolutional lstm network: A machine learning approach for
  precipitation nowcasting.
\newblock In {\em Advances in Neural Information Processing Systems (NeurIPS)},
  2015.

\bibitem{xu2015show}
K.~Xu, J.~Ba, R.~Kiros, K.~Cho, A.~Courville, R.~Salakhudinov, R.~Zemel, and
  Y.~Bengio.
\newblock Show, attend and tell: Neural image caption generation with visual
  attention.
\newblock In {\em International Conference on Machine Learning (ICML)}, 2015.

\bibitem{zhang2015shadow}
L.~Zhang, Q.~Zhang, and C.~Xiao.
\newblock Shadow remover: Image shadow removal based on illumination recovering
  optimization.
\newblock {\em IEEE Transactions on Image Processing (TIP)}, 24(11):4623--4636,
  2015.

\bibitem{zhang2018progressive}
X.~Zhang, T.~Wang, J.~Qi, H.~Lu, and G.~Wang.
\newblock Progressive attention guided recurrent network for salient object
  detection.
\newblock In {\em IEEE Conference on Computer Vision and Pattern Recognition
  (CVPR)}, 2018.

\bibitem{zhu2010learning}
J.~Zhu, K.~G. Samuel, S.~Z. Masood, and M.~F. Tappen.
\newblock Learning to recognize shadows in monochromatic natural images.
\newblock In {\em IEEE Conference on Computer Vision and Pattern Recognition
  (CVPR)}, 2010.

\bibitem{zhu2018bidirectional}
L.~Zhu, Z.~Deng, X.~Hu, C.-W. Fu, X.~Xu, J.~Qin, and P.-A. Heng.
\newblock Bidirectional feature pyramid network with recurrent attention
  residual modules for shadow detection.
\newblock In {\em European Conference on Computer Vision (ECCV)}, 2018.

\end{thebibliography}
}

\end{document}